\documentclass[11pt,a4paper]{article}
\usepackage{times,latexsym}
\usepackage{url}
\usepackage[T1]{fontenc}

\usepackage[acceptedWithA]{tacl2021v1}
\usepackage{tacl2021v1}
\usepackage{xspace,mfirstuc,tabulary}

\newif\iftaclinstructions
\taclinstructionsfalse % AUTHORS: do NOT set this to true
\iftaclinstructions
\renewcommand{\confidential}{}
\renewcommand{\anonsubtext}{(No author info supplied here, for consistency with
TACL-submission anonymization requirements)}
\newcommand{\instr}
\fi

\iftaclpubformat % this "if" is set by the choice of options

\else

\fi

\title{Instructed to Bias: Instruction-Tuned Language Models \\  Exhibit Emergent Cognitive Bias}

\author{Itay Itzhak\textsuperscript{1}, Gabriel Stanovsky\textsuperscript{2}, Nir Rosenfeld\textsuperscript{1}, Yonatan Belinkov\textsuperscript{1} \\
\textsuperscript{1}Technion -- Israel Institute of Technology \\
\textsuperscript{2}School of Computer Science and Engineering, The Hebrew University of Jerusalem \\
\qquad
{{\tt itay1itzhak@gmail.com},} \\
{ \tt \{nirr, belinkov\}@technion.ac.il}, {\tt gabriel.stanovsky@mail.huji.ac.il}
}

\date{}

\usepackage{booktabs}
\usepackage{graphicx}
\usepackage{multirow}
\usepackage{bbm}
\usepackage{bm} % for the \boldsymbol
\usepackage{makecell} % for newline in table cells

\definecolor{bottlegreen}{rgb}{0.0,0.42,0.31}
\definecolor{donorred}{RGB}{228.,116.,95.}
\definecolor{reciepientblue}{RGB}{0,152,251}
\definecolor{torquise}{RGB}{0,206,209}

\usepackage[normalem]{ulem}

\newcommand{\biaseddataset}{treatment}
\newcommand{\Biaseddataset}{Treatment}
\newcommand{\formulabiaseddataset}{T}

\definecolor{my_blue}{RGB}{77,166,255}
\definecolor{my_pink}{RGB}{255,153,238}

\newcommand{\iedit}[1]{{#1}}
\newcommand{\ieditso}[2]{{#2}}

\begin{document}
\maketitle
\begin{abstract}

Recent studies show that instruction tuning (IT) and reinforcement learning from human feedback (RLHF) improve the abilities of large language models (LMs) dramatically.
While these tuning methods can help align models with human objectives and generate high-quality text, not much is known about their potential adverse effects. 
%\nir{i don't think we can/should claim that cognitive-like biases are adverse. maybe humans prefer their machines to also display biases, like them?} \itay{maybe, but these were not the training objectives of these models (and the opposite in some regards as they are intended to be rational following instructions agents.)}
In this work, we investigate the effect of IT and RLHF on decision making and reasoning in LMs, focusing on three cognitive biases%\nir{i don't think it's correct to say `in' cognitive biases. maybe say `in tasks where humans are known to exhibit...'. or just say `focus on', but then it's a bit out of context}\itay{changed to focus, I think it works}
---the decoy effect, the certainty effect, and the belief bias---all of which are known to influence human decision-making and reasoning.
% We cast three well known types of cognitive biases
% This paper provides evidence that fine-tuned models exhibit cognitive-like biases, which were absent or less pronounced in their pretrained predecessors.
%\gabisrep{pretrained counterparts}{predecssors? It's just that these models are also pretrained}.
% We examine the extent of this phenomenon in three cognitive biases---the decoy effect, the certainty effect, and the belief bias---all of which are known to influence human decision-making and reasoning.
Our findings highlight the presence of these biases in various models from the GPT-3, \iedit{Mistral}, and T5 families. Notably, we find a stronger presence of biases in models that have undergone instruction tuning, such as Flan-T5, \iedit{Mistral-Instruct}, GPT3.5, and GPT4. %\gabis{this last sentence felt repetitive, consider omitting} \itay{It's important enough to say that we check different models and not just GPT3}
Our work constitutes a step toward comprehending cognitive biases in instruction-tuned LMs, which is crucial for the development of more reliable and unbiased language models.\footnote{\url{https://github.com/itay1itzhak/InstructedToBias}} 

\end{abstract}
\begin{table*}[th!]
\centering
%\small
%\caption{Examples of Samples From Each Dataset.}
\begin{tabular}{lp{0.41\textwidth}p{0.41\textwidth}}
\hline
\hspace{0.3cm} \textbf{Bias} & \hspace{2.7cm} \textbf{Control} & \hspace{2.3cm} \textbf{\Biaseddataset{}} \\

\midrule
\hspace{0.15cm} Decoy & \makecell[cl]{Below you will find three phone brands.\\ Which one would you choose?\\ Brand 1 - price is \$130, quality rating is 40. \\
Brand 2 - price is \$350, quality rating is 60. \\
Answer: \textbf{\texttt{\textcolor{blue}{Brand 1.}}}} & \makecell[cl]{Below you will find three phone brands.\\ Which one would you choose?\\ Brand 1 - price is \$130, quality rating is 40.\\
Brand 2 - price is \$350, quality rating is 60.\\ 
\textcolor{red}{Brand 3 - price is \$438, quality rating is 60.}\\ 
Answer: \textbf{\texttt{\textcolor{orange}{Brand 2.}}}} \\
\midrule

Certainty & \makecell[cl]{Choose between:\\Option A - \$4000 with a 20\% chance,\\ \$0 with an 80\% chance.\\ Option B - \$3000 with a 25\% chance,\\ \$0 with a 75\% chance.\\ What is your choice?\\Answer:\textbf{ \texttt{\textcolor{blue}{Option A.}}}} &
 \makecell[cl]{Choose between:\\Option A - \$4000 with an 80\% chance,\\ \$0 with a 20\% chance.\\ Option B - \$3000 with \textcolor{red}{certainty}.\\ What is your choice? \\ Answer: \textbf{\texttt{\textcolor{orange}{Option B.}}}} \\
\midrule

\hspace{0.15cm} Belief & 
  \makecell[cl]{Determine if the following argument is\\ logically valid -\\All zint are thade. \\
  Some thade are snaff things. \\
  Conclusion: Some zint are snaff things. \\
  Answer: \textbf{\texttt{\textcolor{blue}{This argument is }}} \\ \textbf{\texttt{\textcolor{blue}{invalid.}}}} &
  \makecell[cl]{Determine if the following argument is\\ logically valid -\\All diamonds are gems. \\
  Some gems are transparent things. \\
  Conclusion: \textcolor{red}{Some diamonds are }\\ \textcolor{red}{transparent things.} \\
  Answer: \textbf{\texttt{\textcolor{orange}{This argument is valid.}}}}  \\ 
\bottomrule

\end{tabular}
\caption{
Illustrative examples of the three evaluated Biases. \textcolor{red}{Red text} indicates disruptive elements fueling the bias. \textcolor{blue}{Blue text} represents control responses unhindered by bias, while \textcolor{orange}{orange text} denotes treatment responses influenced by bias. The \textbf{decoy effect} in the first row presents a scenario where two prize options are compared, the \textbf{certainty effect} in the second row involves selecting products with varying prices and quality measurements, and the \textbf{belief bias} in the third row entails evaluating the validity of logical syllogisms. In the certainty effect and decoy Effect, the model is tasked with choosing its preferred option, whereas in the belief bias, the model determines the conclusion's validity. Each bias is evaluated using a control and a \biaseddataset{} datasets. A shift in choice patterns is anticipated from model predictions on samples transitioning from the control dataset to the \biaseddataset{}. %\gabis{I think that this is an important table, and can benefit from some formmating, e.g., color or markups to emphasize what's different between control and treatment for example}
}
\label{table:examples_biases}
\end{table*}

\section{Introduction}

Advanced fine-tuning methods, like instruction tuning (IT) and reinforcement learning from human feedback (RLHF), have recently emerged as essential paradigms for improving the alignment of language models (LMs) with human objectives \cite{Ouyang2022TrainingLM,bai2022training}.
Although widely adopted \cite{zhou2023comprehensive}, the specific cases in which IT and RLHF enhance model behavior to resemble human behavior, and the mechanisms involved in this process, remain unclear.

In this work, we delve into the impact of IT and RLHF on decision-making and reasoning in LMs.
Recent studies highlighted to some extent cognitive-like biases in pretrained LMs \cite{Binz2022UsingCP, dasgupta2022language} and instruction-tuned models \cite{hagendorff2022machine}.
We take a step further, exploring the consequences of IT and RLHF interventions on LMs' cognitive-like behavior.
We inspect three well-researched and fundamental biases:
the decoy effect
%\cite{Huber1981AddingAD,wu2020profiting}, 
\cite{huber1982adding}, 
the certainty effect %\cite{kahneman1979prospect,schmidt1998measurement}, 
\cite{kahneman1979prospect}, 
and belief bias
%\cite{evans1983conflict,evans1995belief,klauer2000belief}, 
\cite{evans1983conflict}.
% have virtually transformed much of the traditional economic frameworks and
These biases reflect basic inconsistencies in human decision-making (decoy and certainty effects)
and fallacies in logical reasoning (belief bias)
that are prevalent, persistent, and consequential
\cite{berthet2022impact,acciarini2021cognitive}.
% \nir{do these citations refer to `everyday life', or to lab studies?} \itay{changed to a more general phrase.}

The conventional approach to studying cognitive biases in humans is to design simple experiments that elicit from human subjects either judgments or decisions that are likely to reflect a target bias.
Many of these experiments involve question answering;
%Table \ref{table:examples_biases}
Table \ref{table:examples_biases} shows examples of questions used in such experiments,
illustrating how the responses of subjects can suggest biased behavior.
%We test our hypothesis by adapting classic cognitive experiments and incorporating custom data into the LM settings.
To study cognitive-like biases in LMs, %our approach relies on 
% We conduct our investigation by 
we 
adapt classic human experiments to an LM setting.
Towards this, we create an experimental dataset using semi-automatic generated decision tasks:
% This data generation involves
%leveraging the settings of cognitive experiments,
First, for each bias, we manually create an array of appropriate task templates containing flexible numeric and textual placeholder variables.
Then, for a range of values and sets of alternatives, we generate a large collection of unique textual prompts, which we then use as queries to LMs.
%designed to capture the patterns of the biases.
% Like the classic experiments, our semi-automatic generated data is divided into control and treatment data sets.
% We then measure the bias quantity as the difference between the models' choices in the two data sets. 
Following the classic experimental paradigm,
in each experiment we partition the generated data into a `control' dataset and a 'treatment' dataset, and define and measure the bias of a given LM as the average difference of its choices between the two datasets.

Within this setup, we empirically evaluate the degree of bias exhibited by several pretrained LMs, and compare them to their corresponding fine-tuned variants.
Our findings indicate that applying IT or RLHF tuning either
\emph{introduces} cognitive-like biases into text generation,
or \emph{amplifies} these biases if they already exist.
%For example, Figure \ref{fig:figure1_example} presents an evaluation of the certainty effect on the GPT3 and GPT3.5 models, showing that the IT-tuned LMs present a bias that was not found in the pretrained LM.

%\nir{this seems incomplete - what is this sentence trying to convey?}
%\gabist{Our findings present compelling evidence that instruction-tuned and RLHF-tuned models display biases that were either absent or less prominent in their pretrained counterparts.} \gabis{I don't like superlatives, let the readers decide if it's compelling.} \itay{sure, so just erase `compelling', not the whole sentence, right?}

Given that fine-tuned models are typically considered to be superior,
our results point to an important limitation of tuning based on instructions or human feedback.
% namely their tendency to display cognitive-like biases.
% Additionally, the human feedback-tuned models
%\gabis{by latter we mean RLHF or also finetuned}
Fine-tuned models are also often regarded as potentially \emph{less biased},
such as in domains like gender or race,
since they can be explicitly trained to avoid these biases or having personal preferences 
\cite{2303.08774}.
% However, like previous efforts to debias models \cite{gonen2019lipstick}, 
% % \nir{so instruction/RLHF is also a debiasing effort? i thought you guys said it wasn't...} \itay{instruction tuning is not, RLHF could be used to debias and was used in GPT models.}
% when faced with biases these models were not explicitly trained to handle, these models may reveal the limits of these training methods.
% \nir{`faced with biases' seems strange to me. but i am in favor of saying something along the lines of `fine tuning tries to solves one problem, but causes another one to pop up'}
%\gabis{I don't see the connection to Hila's work. Was RLHF specifically designed to make models less biased?} \itay{according to OpenAI, yes, among other reasons. They specifically show in the referred report how RLHF 'fixed' biased behavior.} \gabis{so it's worth mentioning that IMO, I don't know if that's well known}
Our results suggest that, 
similarly to debiasing attempts \cite{gonen2019lipstick},
improving alignment with respect to one human objective
may result in behavior that is unintended with respect to others.
%\nir{rephrased this - please make sure it makes sense} \itay{changed it to be more accurate}

% perhaps attempts of model tuning using human feedback to avoid or adopt explicit human-like preferences may inadvertently give rise to new implicit preferences.
% \nir{i would be very careful in how we phrase this paragraph - this is a strong claim to make, and so we should be very precise} \itay{changed it to be softer. Anyway it should be OK since the rise of some implicit preferences has already been shown in non academic work (like the preference for longer text which is known and talked about)}
\section{Cognitive Biases: Background and Experimental Setup}\label{sec:cog_biases}

Rational choice theory depicts humans as making choices in a manner that maximizes value
on the basis of fixed preferences.
% Rational choices reflect people's fixed preferences that guide decision-making and reasoning processes.
A large body of literature is devoted to describing how actual human behavior deviates from this ideal.
Cognitive biases aim to explain regular inconsistencies in choice behavior
by revealing our susceptibility to `supposedly irrelevant' factors,
such as the context of the decision task, or its framing.
Cognitive biases are therefore defined and measured by how judgments and decisions
deviate from the rational or logical ideal in response to contextual changes.
% , which, from a rational or logical perspective, should not have a significant impact.

%\paragraph{Why We Chose These Biases.}
 Our research targets three biases
that are both prevalent and well-established.
The first two are the \emph{decoy effect} and the \emph{certainty effect} --- decision-making biases that relate to special cases of the more general \emph{prospect theory} \cite{kahneman1979prospect},
with each capturing one of its distinct aspects: the perception of value, and the perception of uncertainty.
%\nir{if you want, you can say that together both biases consider that two basic elements of prospect theory - how people perceive (or act on the perception of) (i) value, and (ii) probabilities} \itay{I prefer to keep it concise here.}
The third bias is the \emph{belief bias},  a logical fallacy in judgment, 
which was previously observed in a closed-off pretrained model \cite{dasgupta2022language}.%\nir{what do you mean by `solely' here?}\itay{delete `solely', didn't add meaning.}
%\nir{i didn't understand this last statement} \itay{rephrased it.}

In this section, we provide for each bias some general background and a description of its classic experimental setup, which we later build on.

%Additional information on the definition of cognitive bias and why we chose these biases can be found in Appendix \ref{appendix:cog_bias_extra_info}. %\gabis{I think that ``why we chose these biases'' should appear here, seems like an important design choice which should be conveyed in the main text. Maybe work with Nir on a sentence or two motivating our choice?}

%\nir{add transition sentence (``we now describe each bias...'')?}
%We now proceed to delve into the detailed description of each bias.

% \nir{perhaps write some more on cognitive biases in general, and ours in particular? give some history, context, describe their scope and prevalence, etc}
% \itay{I suspect it would be too much for the average NLP reader. We can refer to a survey maybe?}

\subsection{Decoy Effect} \label{subsec:def_decoy}
% \nir{i suggest the following structure for the `background' pars (for all three biases): start with setup, and what rational choice predicts (`when given a set of alternatives to choose from, a rational agent would choose her most preferred alternative, i.e., the one having the highest intrinsic value'); then say how general deviations from this look like (`but in reality, human choices are often affected by the set of alternatives presented to them'); then drill in to the particular effect; (`for example, in choosing between A and B, the existence of a clearly inferior alternative C can affect the choice of A vs. B'); then describe the experimental technique to elicit this sort of behavior (`to study this, experiments use a `decoy' option...');
% then say what the bias is (`when affected by the existence of a dummy option, choice behavior is said to exhibit the `decoy bias' effect').}

\paragraph{Background.} 
When choosing from a set of alternatives, a rational agent chooses the item having the highest intrinsic value.
% Consider a setting in which an agent must choose one item from a set of alternatives.
% individuals often assess multiple options based on various choice criteria.
% Whereas rational agents make choices on the basis of intrinsic item value,
Human choices, however, are often affected by context, and in particular,
by the set of available alternatives.
For example, a decision maker who chooses $A$ from the set $\{A, B\}$
may decide to choose $B$ from the set $\{A,B,C\}$ -- a behavior which cannot be consistent with any underlying preference ordering \citep{mcfadden1974conditional}.\footnote{A rational agent would necessarily choose either $A$ or $C$.}
The extreme case in which $C$ is clearly inferior to both $A$ and $B$,
has been coined as the \emph{decoy effect}, to portray $C$ as a `decoy' item
whose only role is to shift the choice from $A$ and $B$.
\looseness=-1

\paragraph{Experimental Setup.}
% \nir{in my mind, these sub-sections should drill in on the experimental details - but at this point the reader should already have a broad idea of the bias and how its treated experimentally} \itay{I'm not sure what is the suggestion here - should the bias be explained better beforehand or the sub-section should drill into more detail? or both? is this assuming the other changes you suggested will be done correctly?}
To study the decoy effect,
we adopt the experimental setup of \citet{huber1982adding},
who proposed to measure 
how the choice between two items changes when a third \emph{asymmetrically dominated} item---the decoy---is added to the choice set.
% The decoy effect materializes when humans are presented with two options, and will tend to have a specific change in preference when also presented with a third decoy option that is asymmetrically dominated \cite{Huber1981AddingAD}.
Items in the experiment are described by their attributes (e.g., quality and price).
In the control condition, subjects are asked to choose one item out of two comparable alternatives;
in the treatment condition, an additional \textit{Decoy Option} is added to the choice set.
The decoy's attributes are set so that it is asymmetrically dominated
(i.e., is worse in all dimensions) by one of the original items, referred to as the \textit{target option}, but not by the other item, referred to as the \textit{competitor option}.
% an asymmetrically dominated item (\textit{Decoy Option}) 
% when it has a lower rating in all dimensions compared to one option (\textit{Target Option}); but, in comparison to the other option, it has a lower rating in some dimensions and higher in others (\textit{Competitor Option}).
Table \ref{table:examples_biases} (first row) provides a concrete example:
Brand 1 and Brand 2 are comparable, whereas Brand 3 (the decoy) is inferior to Brand 2 (target), but not to Brand 1 (competitor).
%\nir{i think it's confusing that you point to an example, which is essentially how an experiment is designed, before you talk about the experimental design}
% The treatment sample has the same options Brand 1 (\textit{Competitor Option}) and Brand 2 (\textit{Target Option}) as the control sample but was added a third Brand 3 (\textit{Decoy Option}).
% Brand 3 is directly inferior compared to Brand 2.

Choice behavior is said to exhibit the `decoy effect' if 
% The decoy bias describes the phenomenon where people
subjects tend to choose Brand 1 in the control condition,
but prefer Brand 2 in the treatment condition.
By design, this means that choices are affected by a supposedly irrelevant factor---the availability of an alternative that in itself will never be chosen,
suggesting that choices are biased.
% This shift of choice happens despite these two options being the exact same options in both samples.

% Note that for any fixed pair of target and competitor items,
% there is a range of possible values for price and quality which the decoy can take.
% \citet{huber1982adding} studies four sub-types of decoys, with attributes relative
% to the target option being:
% higher price bur same quality (\textit{R});
% extremely higher price but same quality (\textit{R*});
% higher price and worse quality (\textit{RF});
% and same price but worse quality (\textit{F}).
% We compare the effects of each sub-type on models and human choices.

\subsection{Certainty Effect} \label{subsec:def_certainty}
% \nir{see prev. comment on the layout of pars: rational choice (here, maximizing expected value); general deviation (make suboptimal decisions due to distorted perception of probabilities, especially small and large); specific setting (extreme case - p=1); etc}

\paragraph{Background.} Most decision settings involve some degree of uncertainty.
% In most situations, choices involve uncertain outcomes that include obtainable values in certain probabilities.
Given a set of alternatives describing possible outcomes and their probability,
utility theory \citep{friedman1948utility}
determines that rational agents will choose the option with the highest expected value.
% According to classic theory \cite{friedman1948utility}, a rational individual  will usually prefer options with a higher expected value, taking into account the values and probability of each outcome. 
Human choice, however, tends to deviate from this standard,
especially when the probabilities to consider are either very small or very large.
%
% However, a deviation from rational thinking would be a sub-optimal choice in terms of expected value.
% This deviation can happen when one option is presented as \textit{certain}, possibly due to over-weighting the probability of the certain choice.
%
% A certain option leads to a preference for options with lower expected value but greater certainty.
% \nir{we can't really say that preferences shift, or that something `lead' to something... again this is trying to explain the underlying mechanism, but we should stick to descriptive statements}
% tend to display a preference for options with guaranteed or certain outcomes, even when the expected value of an alternative option is objectively higher.
The \emph{certainty effect} describes people's tendency to prefer outcomes that occur with certainty to alternatives that yield higher expected value, but include risk.
This effect was initially explored in the seminal work of \citet{kahneman1979prospect},
whose experimental setup we describe next.%\footnote{We explore an additional adjacent effect from the same work and provide details in Appendix \ref{appendix:sec:not_probable}.}

% underweight outcomes
% that are merely probable in comparison with outcomes that are obtained with certainty.

% The following experimental setup was laid out by \citet{kahneman1979prospect} which described the certainty effect for the first time.

\paragraph{Experimental Setup.}
In \citet{kahneman1979prospect}, human subjects were asked to choose between
two `lotteries', each describing a simple distribution over potential monetary rewards
(e.g., 80\% to win \$100 and 20\% to win nothing).
In the control condition, subjects were given two lotteries $A,B$, each having some degree of risk;
in the treatment condition, alternative $B$, having lower expected value,
was modified to provide its original expected value but with a probability one
(i.e., the same expected value but at no risk).

Table \ref{table:examples_biases} (second row) presents an example.
% These experiments are illustrated using the samples in the second row of Table \ref{table:examples_biases}.
% In this example, both the treatment and control samples present two options with varying probabilities and prize gaps.
In both conditions, the prize in Option A remains the same and has a higher expected reward than Option B,
whose certainty varies across conditions.
% The bias represents a tendency to choose Option A with higher expected utility in the control sample, yet favoring the certain Option B in the treatment sample despite its lower expected utility.
As in the example,
 \citet{kahneman1979prospect} (and many follow-up studies)
found that, while control subjects tend to choose rationally,
treatment subjects display a strong preference towards the certain alternative despite its lower expected reward.

\begin{table*}[th!]
\centering
% \small
\begin{tabular}{@{}llcccc  cc@{}}
\toprule
  && \multicolumn{4}{c}{\textbf{Decoy}} & \multicolumn{1}{l}{\textbf{}} & \multicolumn{1}{l}{\textbf{}}  \\
 \cmidrule(lr){3-6}
\hspace{0.2cm} \textbf{Condition}  && Frying Pan & Phone & Car & Real-Estate & \textbf{Certainty}  & \textbf{Belief}  \\
\midrule

\hspace{0.3cm} \textbf{Control}
 & \# Samples& 96 & 120  &  120 &  96 & 336  & 672 \\

\hspace{0.1cm} \textbf{\Biaseddataset}
 & \# Samples& 1152 & 1440  & 1440 &  1152 & 504  & 672 \\
\midrule

 \hspace{0.3cm} \textbf{Templates}& \# Prompts& 4& 4& 4& 4& 3*&7\\
\hspace{0.05cm} \textbf{Values Range}& US-Dollars & 9.99-179.99& 100-900& 5K-35K& 80K-500K& 2.4K-5K&NA\\\bottomrule
\end{tabular}
\caption{\iedit{Sample and template counts in each dataset, along with the range of values for decoy products and certainty effect prizes. The different text templates and values were used to evaluate the biases robustly while using reasonable values and phrasing. The (*) notation for certainty effect templates denotes the primary textual prompt without sub-templates.}}
\label{tab:data_numbers}
\end{table*}

\subsection{Belief Bias} \label{subsec:def_belief}
\paragraph{Background.}
%\nir{here, need to say what part of this is `judgment'}
% Syllogisms are a class of reasoning problems characterized by a straightforward argument structure, involving two true statements that logically necessitate a third statement \cite{smith2000aristotle}. 
% These multi-step reasoning tasks provide an essential framework for investigating deductive reasoning abilities and the cognitive processes involved in drawing valid conclusions.
Syllogisms are a class of reasoning problems involving two true statements and a third conclusion statement, which is either logically deductible from the true statements, or is not \cite{sep-aristotle-logic}. 
To make a rational judgment of the conclusion,
it is both necessary and sufficient to apply logical reasoning
to the true statements---and to them alone. %regardless of other world knowledge. 
\emph{Belief bias} occurs when a person's evaluation of the conclusions' validity is affected also by their own knowledge, beliefs, or values,
which can sometimes lead to false reasoning.
% rather than actual evidence presented by the two true statements.
This bias was empirically demonstrated by \citet{evans1983conflict},
whose results suggest that human judgment can be affected by the
`believability' of the conclusions,
i.e., that subjects' perception of logical validity depends on the degree
to which the conclusion %statement 
is believable (or not).

\paragraph{Experimental Setup.}
In \citet{evans1983conflict}, human subjects were given sets of two premises and a conclusion, and asked whether the conclusion logically followed from the premises \cite{evans1983conflict}.
Half of the conclusions were phrased to be \textit{believable}---aligned with general world knowledge (e.g., ``cigarettes are addictive''), and the other half was constructed to be \textit{non-believable} (``cigarettes are non-addictive'').

Table \ref{table:examples_biases} (third row) shows an example.
Both treatment and control tasks include two premises and an invalid conclusion;
while the control includes fictitious objects,
the treatment includes real-world objects---which in this case are believable, and entail an erroneous answer (`valid').
% , , leading to more relatable conclusions.
% The bias reflects a propensity to invalidate the control sample's conclusion while validating the treatment's conclusion.
%The results of
\citet{evans1983conflict} showed that subjects tended to consider believable conclusions as valid and unbelievable conclusions as invalid,
suggesting the presence of belief bias in their judgments.

% In recent work, \citet{dasgupta2022language} showed that Chinchilla  \citep{hoffmann2022training} exhibits belief-bias like behavior.
% % Our work recreates their results on this bias and expands them to other models.
% Our work investigates more models from different models' families, focusing on the effect of IT and RLHF on the extent of the biases.\nir{is this the place to state this? (didn't we already say something more general before?}

\section{Data and Evaluation}
We next describe our data generation process and evaluation scheme.
Sec.~\ref{subsec:data}, outlines our semi-automatic approach for generating specific datasets, each designed to probe a certain cognitive bias and to evaluate the existence of biased `behavior' in models.
\iedit{Sec.~\ref{subsec:data_overview} provides further details on these generated datasets.}
Sec.~\ref{subsec:bias_score} formally introduces our proposed \emph{bias score}, intended to quantify the degree of bias exhibited by a model based on its predictions on the generated data. %\gabis{change the order of the sentence: ``in Section ..., we quantify ...''}

\begin{table*}[th!]
\centering
\small
\begin{tabular}{@{}llccc  cc@{}cc}
\toprule
 &  & \multicolumn{3}{c}{\textbf{GPT3}} & \multicolumn{2}{c}{\textbf{T5}} & \multicolumn{2}{c}{\textbf{Mistral}}\\
 \cmidrule(lr){3-5} \cmidrule(lr){6-7} \cmidrule(lr){8-9}
& & \textbf{LM} &  \multicolumn{2}{c}{\textbf{IT-LM}} & \textbf{LM} & \textbf{IT-LM}  &\hspace{0.5cm}  \textbf{LM} &\textbf{IT-LM}  \\
\cmidrule(lr){4-5}
& \hspace{0.8cm} \textbf{Bias} & DaVinci & DaVinci-002 & DaVinci-003 & T5 & Flan--T5    &\hspace{0.25cm} Mistral&Mistral-I\\
\midrule
\multirow{5}{*}{\rotatebox[origin=c]{90}{\textbf{Bias Score}}} 
& Decoy Expensive & -- 0.15* & -- 0.13* & \textbf{-- 0.02} & \textbf{0.02} & -- 0.18*     & 0.03 &\hspace{0.08cm} \textbf{0.24}**\\
& Decoy Cheaper & -- 0.17* & -- 0.08* & \textbf{0.08}** & -- 0.15* & \textbf{0.20}*   & -- 0.05* & -- \textbf{0.03}**\\
%& Certainty & 0.00 & 0.21* & \textbf{0.54}* & 0.09* & \textbf{0.16}*  \\ % this is 3 sub-types
& Certainty & 0.00 & 0.24* & \textbf{0.67}* & 0.09* & \textbf{0.17}*    &0.03&\textbf{0.29}*\\ % this is two main sub-types
& Belief Valid & 0.00 & 0.19* & \textbf{0.21}* & -- 0.03 &\textbf{0.50}*      & 
 0.01 & \textbf{0.26}*
\\
% \midrule
& Belief Invalid & 0.04 & 0.55* & \textbf{0.65}* & 0.03 & \textbf{0.39}*      & 0.05 &\textbf{0.31}*\\
\bottomrule
\end{tabular}
\caption{The difference between the choices of models in the target option under the \biaseddataset{} condition versus the control condition. A higher score means the model exhibits a higher level of bias. In bold are the highest values in each model family. (*) Marks results that are statistically significant with p-values $<.05$,  and (**) marks results that are averaged across multiple products where some are significant and others are not. \iedit{Mistral-I refers to Mistral-Instruct.}}
\label{tab:results}
\end{table*}

\subsection{Data Generation} %\gabis{use title headers more descriptively, what kind of data? Think of a reader who's skimming the paper and would like to decide if they want to read this subsection}
\label{subsec:data}
% To evaluate the extent to which a model is biased, for each bias, we generate a biased dataset and a corresponding control unbiased dataset and compare them.
To assess the level of each bias in a model, we employ a comparative approach, as shown in Table \ref{table:examples_biases}.
We do that by
%\gabis{comparing a generated ... (so the reader can link this back to the ``comparative'' method described above)}
comparing predictions on a generated \biaseddataset{} dataset and a corresponding control dataset.
%\gabis{In contrast? Why are we singling out belief bias? Is the method different from the other two? If so, let's say it explicitly}
For the decoy and certainty effects, we use data generated according to values crafted by us, and in the belief bias we use data generated in a similar fashion by \citet{dasgupta2022language} with additional text templates we wrote. 

 To generate the \biaseddataset{} datasets, we follow the experimental design for each bias as outlined in Section \ref{sec:cog_biases} \iedit{and use new values that align with the cognitive experiments methods}.

The control versions of the datasets are carefully crafted to closely resemble the \biaseddataset{} samples while excluding the specific attribute that triggers the bias, as identified by cognitive experiments.

In the decoy and certainty effects, for each sample, there exists a designated \textit{Target} option.
This option is expected to be chosen more frequently by a human (or a biased model) when presented with samples from the \biaseddataset{} dataset compared to samples from the control dataset.
In the belief biases, we treat the correct answer as the \textit{Target} option, for ease of notation.
We later use the \textit{Target} option to compute the bias scores, as detailed in Section \ref{subsec:bias_score}.
%More details about the data generation process can be found in Appendix \ref{sec:appendix_data_generation}.

\subsection{Data Overview}
\label{subsec:data_overview}
% \gabis{I'm not sure we need the paragraph format here instead of just running text} \itay{deleted the paragraphs}

\iedit{Table \ref{tab:data_numbers} provides quantitative metadata for the datasets. We elaborate below on the text templates and values chosen for each bias dataset according to cognitive theory as outlined in Section \ref{sec:cog_biases}.}

%\paragraph{Text Templates.}

\iedit{We used 3, 4, and 7 prompt templates for the certainty effect, decoy effect, and belief bias, respectively.
The certainty effect featured extra sub-templates with variations in option presentations like probabilities or percentages.
All possible permutations of option orders were used for decoy and certainty effects, as well as for both premises in belief bias.
%No significant differences in results were observed among the templates for any of the biases.
%\gabis{is this the right place to talk about results?}
}

%\paragraph{Decoy Values.}

\iedit{Regarding the decoy effect, we utilized realistic values from US-based store websites to construct our datasets.
Quality ratings ranged from 60 to 90 with 10-20 intervals between options.
Decoy options, in comparison to the target, exhibit a 25\% or 50\% price change, a 10-20 point quality rating shift, or a combination of both.
Modern alternatives to the original products were anecdotally chosen, emphasizing a one-time, deliberate selection process without trial and error.}

%\paragraph{Certainty Values.}

\iedit{In line with cognitive bias theory, we chose certainty effect prizes and probabilities to closely mirror the cognitive research data, ensuring accurate expected utility differences between the options.}

%\paragraph{Belief Values.}

\iedit{Belief bias samples involve manually composing both believable and unbelievable arguments, derived from previous work.
The samples are evenly split, with half being believable and the remaining half being unbelievable.
The samples' arguments are built upon simple, well-known objects, such as `All guns are weapons' and `All lizards are reptiles'.
Further details can be found at \citet{dasgupta2022language}}

% \nir{i think at this point the reader should have a clear idea of our methodology. so i suggests bringing forwards some things that appear later: a general template of the tasks we run (ie prompting a classic experiment, letting the model choose one of the available alternatives, measuring outcomes). it should also be clear that we have two conditions (control and treatment) in each task, and that we are interested in comparing them (say how).}

\subsection{Computing The Bias Scores }
\label{subsec:bias_score}
% \gabis{Nit: keep a consistent header scheme across the subsections of the same section. 5.1 has an active verb (``determining''), this one is a nominal phrase (``computation''), and the last one doesn't have a verb. I'd make all of them either active or nominal.}

% We evaluate the presence of biases in each model by examining the differences in their prediction patterns between biased and unbiased datasets.
% We quantify these biases by computing bias scores.

% For the decoy and certainty effects, the bias score is computed as the difference between the model's preference for the \textit{Target} option in the biased condition and the unbiased condition.
% For instance, if the model selected the \textit{Target} option in 90\% of the samples in the biased condition and in 70\% of the samples in the unbiased condition, the bias score would be $0.90 - 0.70 = 0.20$.

We assess biases in each model by analyzing their prediction patterns across \biaseddataset{} and control datasets, quantifying them through bias scores.
The bias score captures the difference in the model's inclination towards the \textit{Target} option between \biaseddataset{} and control scenarios.
%Similarly, in the belief bias, the bias score indicates the model's preference for the \textit{Target} option, which corresponds to the correct answer (valid or invalid).

For example, if the model chose the `Target' option in 90\% of \biaseddataset{} samples and 70\% of control samples, the bias score would be $0.20$.

\paragraph{Bias Score Definition.}
The bias score is formally defined in Equation \ref{equation:bias_score}, where \textit{\Biaseddataset{}} and \textit{Control} represent the sets of \biaseddataset{} and control datasets, respectively, and \textit{$N_{\formulabiaseddataset}$} and \textit{$N_{C}$} indicate their respective set sizes.
$Ans_{i}$ denotes the model's choice in sample $i$, while $T$ represents the target option.

\begin{equation} \label{equation:bias_score}
\resizebox{.85\hsize}{!}{$\sum\limits_{i \in \Biaseddataset{}} \frac{{\mathbbm{1}{ [Ans_{i} \boldsymbol{=} T] }}}{N_{\formulabiaseddataset{}}}
 \boldsymbol{-} \sum\limits_{i \in Control} \frac{{\mathbbm{1}{ [Ans_{i} \boldsymbol{=} T] }}}{N_{C}} $}
\end{equation}

% \paragraph{Bias Score for Random Choice} \gabis{Use the paragraph environment instead of textbf. But also, why is this bolded anyway?}
% When examining the bias score, a value of 0 indicates the absence of bias.
% However, it is noteworthy to consider the decoy effect scenario where the model's selection is randomized. In this case, the bias score would be -0.17.
% This is because, in the unbiased condition, a random choice between the two options would result in selecting the target option 50\% of the time, while in the biased condition with three options, a random choice would lead to the model selecting the target option only 33\% of the time.
% Consequently, this discrepancy yields a bias score of -0.17.

According to the original experimental setting of the decoy effect, the target option \iedit{in each sample} can be associated with either a lower or higher price, leading to the computation of separate bias scores: \textit{Decoy Cheaper} and \textit{Decoy Expensive}.

To compute bias scores for the belief bias, we compare the model's predictions between consistent and inconsistent conditions for valid and invalid arguments. 
This analysis results in two distinct bias scores \iedit{that were recognized in the original experiments}:

\textit{Belief Valid}: The difference between the model's predictions of consistent valid arguments (valid and believable conclusions in real-life objects condition) and neutral valid conclusions (all valid arguments in non-real object conditions).

\textit{Belief Invalid}: The difference between the model's predictions of consistent invalid arguments (invalid and unbelievable arguments in real-life objects condition) and neutral invalid arguments (all invalid arguments in non-real object scenarios).

\paragraph{The Meaning of Bias Score Values.}
Higher bias score values indicate a greater degree of bias in the model.
The bias scores range from $-1$ to $1$, reflecting the extent \iedit{of the bias} and \iedit{its} direction relative to human biases \iedit{according to cognitive theory}.
\iedit{While the original experiments on human evaluation did not calculate bias scores, the intended alignment of these bias scores is with the strength of bias as per the cognitive theory on human biases.}
A score of $1$ represents maximum bias aligned with human biases, $0$ indicates no bias, and $-1$ denotes maximum bias in the opposite direction to human biases.

The significance of each bias score is measured using the student's t-test \cite{student1908probable}.
%We measure the significance of differences of difference between the bias scores of different models via the interaction of the models and the bias scores using a linear regression model and report the results in Appendix \ref{appendix:sec:diff_between_models}.

\section{Experiments}

\paragraph{Models}
We conduct our experiments on two LM sets.
The first set is pretrained models -- GPT3 `DaVinci' \cite{NEURIPS2020_1457c0d6},  and the publicly available \iedit{Mistral-7B \cite{jiang2023mistral} and} T5 \cite{raffel2020exploring}.\footnote{We use version T51.1: \\  \url{github.com/google-research/text-to-text-transfer-transformer/blob/main/released_checkpoints.md}}
The second set consists of improved versions of the preatrained models fine-tuned using IT and human feedback.
For GPT3, we experiment with GPT3.5 models----`text-DaVinci-002'  and `text-DaVinci-003' (`Davinci-002' and `Davinci-003' for short) \cite{Ouyang2022TrainingLM}---as IT and IT+RLHF models respectively.
\iedit{For the Mistral 7B, we use Mistral 7B-Instruct with the recommended chat template\footnote{\iedit{We used ``You are a helpful assistant. Answer shortly with only your choice with no explanation.'' as the opening instruction.}}.}
For T5 we use the Flan-T5 models \cite{chung2022scaling} as the IT version.
Our primary findings are based on the XXL variant of the T5 models (11B parameters), and we also experiment with the XL variant (3B parameters) to investigate the influence of model size.

%LMs are being constantly improved with goals such as being more user-friendly in text generation, reducing hallucinations, and reducing bias.
Finally, we also experiment with one of the latest commercially available models, GPT4  \cite{2303.08774}, which is considered a state-of-the-art generative model.\footnote{We used the `gpt-4-0314' version with the content ``You are a helpful assistant.''}
However, we do not have access to its pretrained version as it was not publicly released.
We, therefore, use GPT4 only as a reference for a newer model. 

%\subsection{Determining the Model's Answer}
\paragraph{Determining the Model's Answer.}
\label{subsec:eval_ans}
% \gabis{On the whole on the one hand this seems like a small implementation issue, otoh, I couldn't understand what we're actually doing. So we don't do this for the instruction-tuned models? only for the non-instruction?}

Given a prompt asking for a choice, the instruction-tuned models \iedit{using greedy decoding} usually generate text describing their choice, simply as  ``Option 1'' or ``Brand 2''.\footnote{In the certainty effect  $<$5\% of the predictions made by Flan-T5-XXL were not clear and we excluded these examples.}

To assess the pretrained performance for each task, we use the common practice \cite{NEURIPS2020_1457c0d6} of evaluating the likelihood of various candidate answers from a predefined set of possible answers. 
\iedit{This helps prevent models from persistently asking questions instead of providing direct answers, as observed in our initial experiments.}
This evaluation might be affected by a preference of the model to an answer given the context (e.g., given ``Answer:'' the model might give a higher baseline probability to ``Option 2'').
We apply the DC-PMI correction \cite{holtzman-etal-2021-surface} that mitigates this issue by normalizing each answer likelihood within the context of the prompt, relative to a baseline prompt (``Answer:'', in our case).\footnote{Small-scale experiments with DC-PMI correction for the instruction-tuned models led to similar results to evaluation without correction, so we only report the latter.}
% We experimented with a similar evaluation for the instruction-tuned models and got similar results on a smaller scale, so we report only the evaluation based on the generated text.

\paragraph{Using Zero-shot}
\label{paragraph:zero_shot}
The samples used for the decoy and certainty effects are choice-dependent questions with no ``correct'' answer (recall the examples in Table 
\ref{table:examples_biases}).
It is therefore not obvious how to construct few-shot examples, which presumably should have correct labels in the prompt.
Given that we focus on decision inclinations, the zero-shot setup aligns naturally with our investigation of all biases.
Most experiments, unless specified, are in the zero-shot format and involve a single question followed by "Answer:" without extra examples, as shown in Table \ref{table:examples_biases}.

\paragraph{Using Few-shot}
\label{paragraph:few_shot}
Despite the above-mentioned problem, we experiment with an approach that constructs a few-shot setting using samples outside of our data.
We build upon a recent work that suggested that giving few-shot samples without the correct labels could improve model performance by introducing the model with the overall format of the samples \cite{min2022rethinking}.
We detail further and report results in Section \ref{subsec:few_shot_results}.

\section{Results} 
 Table \ref{tab:results} summarizes the bias scores of pre-trained models and their instruction-tuned and RLHF-tuned counterparts.
 We discuss the main takeaways in this section and provide several fine-grained analyses in the next one. 
% The bias scores of models fine-tuned on instructions and human feedback are presented in Table \ref{tab:results}, demonstrating higher levels of bias compared to the pretrained models, which exhibit minimal or negligible bias.

\paragraph{Models fine-tuned using IT and RLHF show a higher bias than their pretrained counterparts.}
Our findings reveal that the models fine-tuned on instructions and RLHF mostly exhibit significantly higher levels of bias compared to their pretrained counterparts, as demonstrated in Table \ref{tab:results}.
While the pretrained LMs demonstrate minimal to no bias, the fine-tuned models display pronounced biases across most categories.
This is evident in the certainty effect row, where the DaVinci, T5 \iedit{and Mistral} pretrained models exhibit bias scores of $0.00$, $0.09$\iedit{, and $0.03$}, respectively.
In contrast, the fine-tuned models display higher bias scores of $0.24$, $0.67$, $0.17$\iedit{, and $0.29$}.
This unexpected result suggests that the fine-tuning process, intended to enhance model performance, inadvertently introduces biases into the decision-making process.

We measured the significance of the differences between models using the difference-in-differences method \cite{dimick2014methods}.
All differences were significant except for DaVinci and DaVinci-002 in the belief valid and in decoy expensive, T5 and Flan-T5 in the certainty effect, and Mistral and Mistral-Instruct in the decoy cheaper.
%Details on the significance of differences between models are in Appendix \ref{appendix:sec:diff_between_models}.

\begin{table}[t!]
\centering
%\small
\begin{tabular}{@{}lccc}
\toprule
& \textbf{Bias}  & \textbf{DaVinci-003} & \textbf{GPT4} \\
\midrule
\multirow{5}{*}{\rotatebox[origin=c]{90}{\textbf{Bias Score}}} 
& Decoy Expensive & 0.00 & \textbf{0.38}*  \\
& Decoy Cheaper   & 0.03  & \textbf{0.05}  \\
%& Certainty  &   \textbf{0.35}  & 0.15*  \\ % all 3 biases
& Certainty  &   \textbf{0.43}*  & 0.20*  \\ % two bias types
& Belief Valid &  \textbf{0.20}* & 0.15*  \\
& Belief Invalid &  \textbf{0.47}*  & 0.41*  \\
\bottomrule
\end{tabular}
\caption{comparison of the results between GPT4 and the most recent GPT3.5 release DaVinci-003 in 1-shot format. Scores marked with * are statistically significant with p-values $<.05$}
\label{tab:gpt4_results}
\end{table}

% \begin{table*}[th]
% \centering
% \begin{tabular}{@{}lcc|cc|cc}
% \toprule
%  & \multirow{2}{*}{\textbf{Bias}} & \textbf{Text-DaVinci-003} & \textbf{GPT4 0--shot} & \textbf{Undecided} & \textbf{GPT4 1--shot} & \textbf{Undecided} \\
% \midrule
% \multirow{4}{*}{\rotatebox[origin=c]{90}{\textbf{Bias Score}}} 
% & Decoy Expensive &  --0.05 & \textbf{0.42}* & 0\% & 0.41* & 0\%  \\
% & Decoy Cheaper &  0.07 & \textbf{0.25}*  & 0\% & 0.12* & 0\% \\
% & Certainty & \textbf{0.54}* & 0.17*  & 18\% & 0.15* & 0\% \\
% & Belief Valid & \textbf{0.23}* & 0.07*  & 2\% & 0.06* & 0\% \\
% % \midrule
% & Belief Invalid &  \textbf{0.61}* & 0.49*   & 0\% & 0.28* & 0\%  \\
% \bottomrule
% \end{tabular}
% \caption{comparison of the results between GPT4 and the most recent GPT3.5 release 'Text-DaVinci-003'. Undecided columns are the percentage of examples for which the model did not output a clear answer. Results marked with * are statistically significant with p-values $<.05$}
% \label{tab:gpt4_results}
% \end{table*}

\paragraph{LMs exhibit biases that align with \ieditso{biases observed in humans}{human biases theory}.}

Intriguingly, our investigation reveals a convergence between the decision-making biases observed in the models and the well-established \iedit{theory on} irrational biases inherent in human decision-making processes.
Recall from Section \ref{subsec:bias_score} that positive values indicate alignment between bias scores and human biases. 
Indeed, tuning using instructions or human preferences generally makes bias scores increasingly 
high.
The negative bias score exhibited by DaVinci in the decoy biases can be explained by choice criteria which, unlike humans, are not value-depended.
In this exceptional case, the model chose the last option offered almost all the time, regardless of the options' content, making its choice more focused on positional preferences.
%\footnote{The negative bias score exhibited by the pre-trained DaVinci in the decoy biases can be explained by positional preferences, as discussed in Appendix \ref{appendix:sec:options_locations}.}
% The presence of these biases in the models provides compelling evidence of their replication and propagation within AI systems.

% This discovery emphasizes the profound impact of fine-tuning on bias amplification among biases that may be unknown to exist in models at the moment, underscoring the urgent need for effective bias mitigation strategies in natural language processing systems.
% This similarity between human biases and model biases highlights the potential connection of implicit biased training data on model behavior, as it induces the models to replicate the inherent biases ingrained in human decision-making processes and other human behaviors. 
% yonatan: This is too verbose and bombastic to my taste. I suggest keeping it short and to the point, and potentially discussing more in the conclusion.

This finding emphasizes the role of fine-tuning on bias amplification on previously undiscovered biases.
In addition, the similarity between \iedit{the theory on} human biases and model biases highlights the potential connection of inherent biases ingrained in human decision-making processes to tuning methods that induce the models to replicate human behaviors. 

\paragraph{IT Amplifies Biases.}

The discernible impact of fine-tuning with IT becomes evident upon comparing the T5 \iedit{versus the} Flan-T5 models \iedit{ and the Mistral versus the Mistral-Instruct models}.
While DaVinci and DaVinci-002 versions may differ by more than IT (exact details are not public), the transparent elucidation of the Flan-T5 fine-tuning process \iedit{and the sole instruction tuning done to the Mistral-Instruct model} allows us to confidently assert that the sole utilization of IT can indeed engender the emergence of biases.
This finding highlights the influential role of fine-tuning methods in amplifying biases within models, shedding light on the intricate relationship between IT and the manifestation of biases.

% \textbf{Reinforcement learning fine-tuning on human feedback seems to amplify the biases} 
% The main difference between 'Text-DaVinci-002' and 'Text-DaVinci-003' is that the latter was trained with reinforcement learning on human feedback using the PPO algorithm\footnote{According to OpenAI at https://platform.openai.com/docs/model-index-for-researchers}.

% This may imply that apart from the bias amplification that may be caused by the instruction fine-tuning, reinforcement learning by itself can also be the cause of these biases appearance.
%\paragraph{Amplification of Biases through Reinforcement Learning Fine-tuning}
\paragraph{RLHF Amplifies Biases.}

Our findings indicate that the application of reinforcement learning fine-tuning from human feedback has the potential to amplify biases within language models further.
This is evident when comparing the DaVinci-002 and DaVinci-003 models, with the latter incorporating reinforcement learning techniques.\footnote{According to OpenAI at \url{https://platform.openai.com/docs/model-index-for-researchers}.}
Notably, while IT may contribute to bias amplification, our results suggest that reinforcement learning, as an independent factor, can also play a significant role in the emergence of these biases.
This observation highlights the complex interplay between reinforcement learning methodologies and the manifestation of biases.

\begin{figure}[t!]
\centering
\includegraphics[width=1\columnwidth]{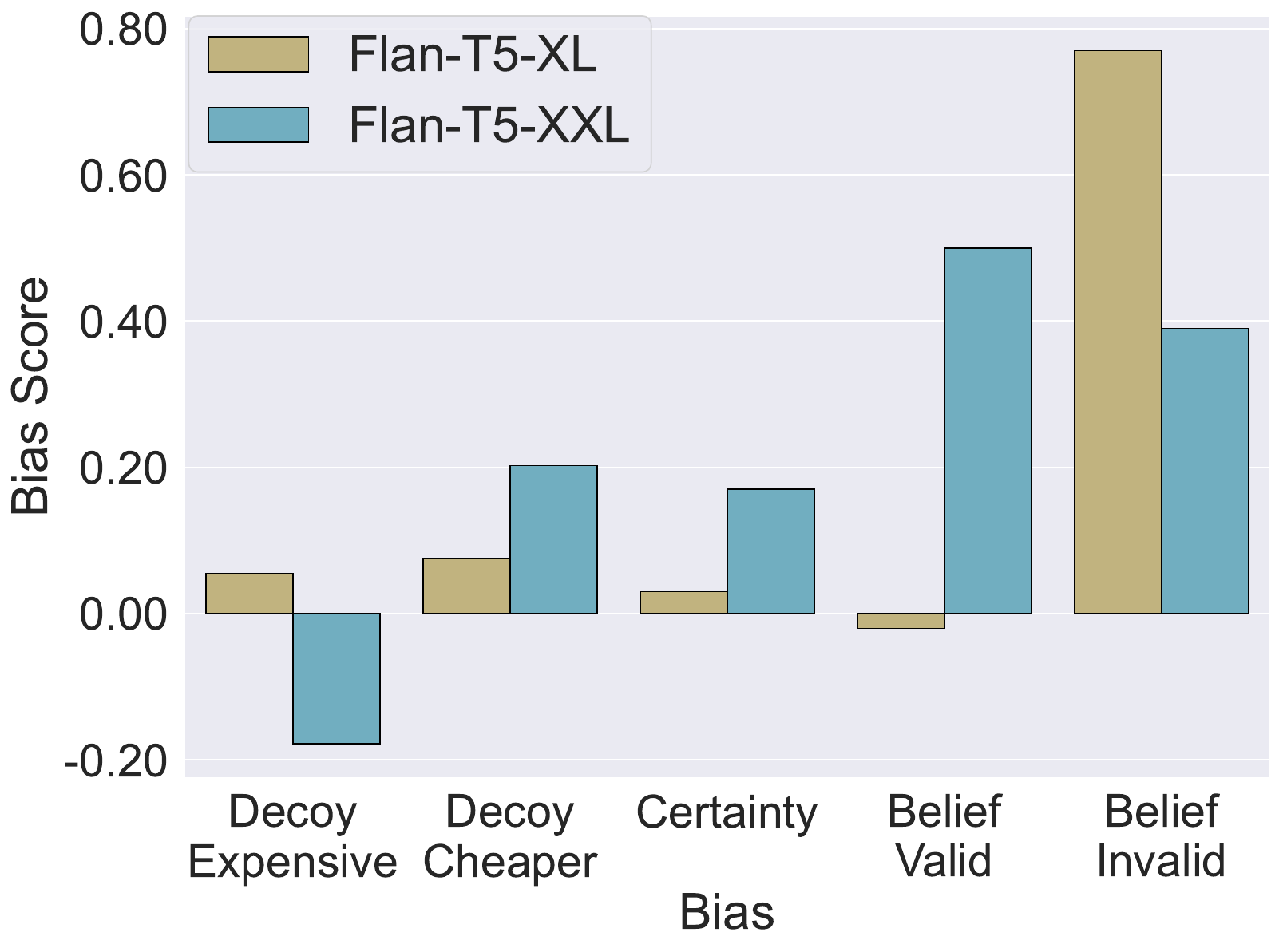}
% \qquad\qquad
\caption{
The impact of model size on bias scores. The larger Flan-T5-XXL exhibits higher bias scores in decoy cheaper, certainty, and belief valid biases while demonstrating lower bias scores in decoy expensive and belief invalid biases compared to the smaller Flan-T5-XL. The decoy expensive bias discrepancy may stem from Flan-T5-XXL's preference for higher-priced products, while the belief invalid bias reduction can be attributed to the model's enhanced accuracy with neutral arguments.}
\label{fig:model_size}
\end{figure}

\begin{figure}[th!]
\centering
\includegraphics[width=1\columnwidth]{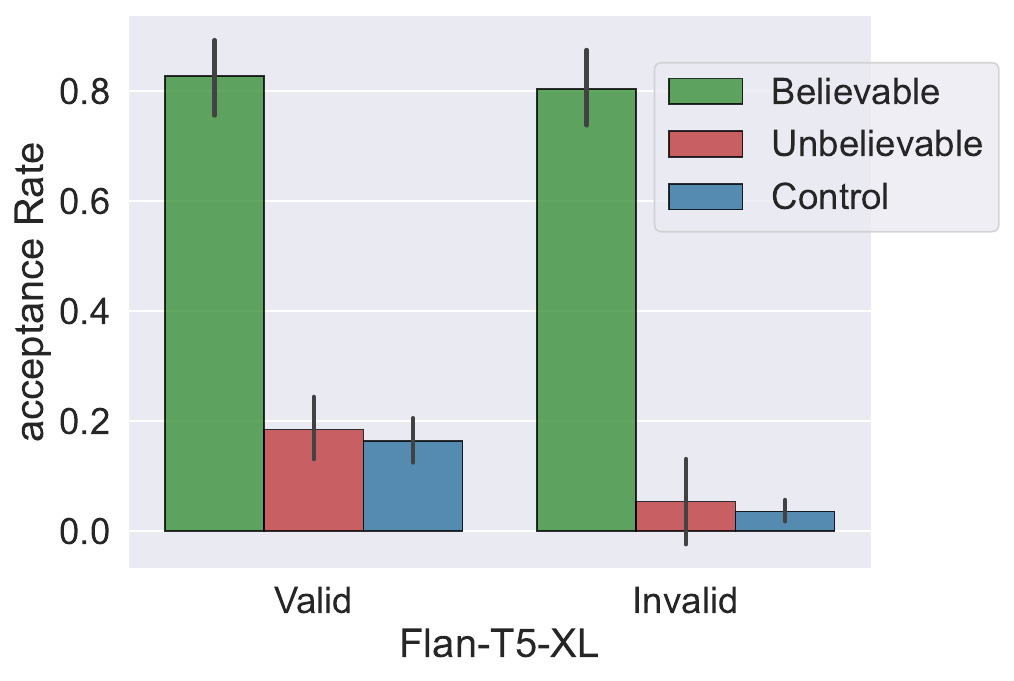}
\includegraphics[width=1\columnwidth]{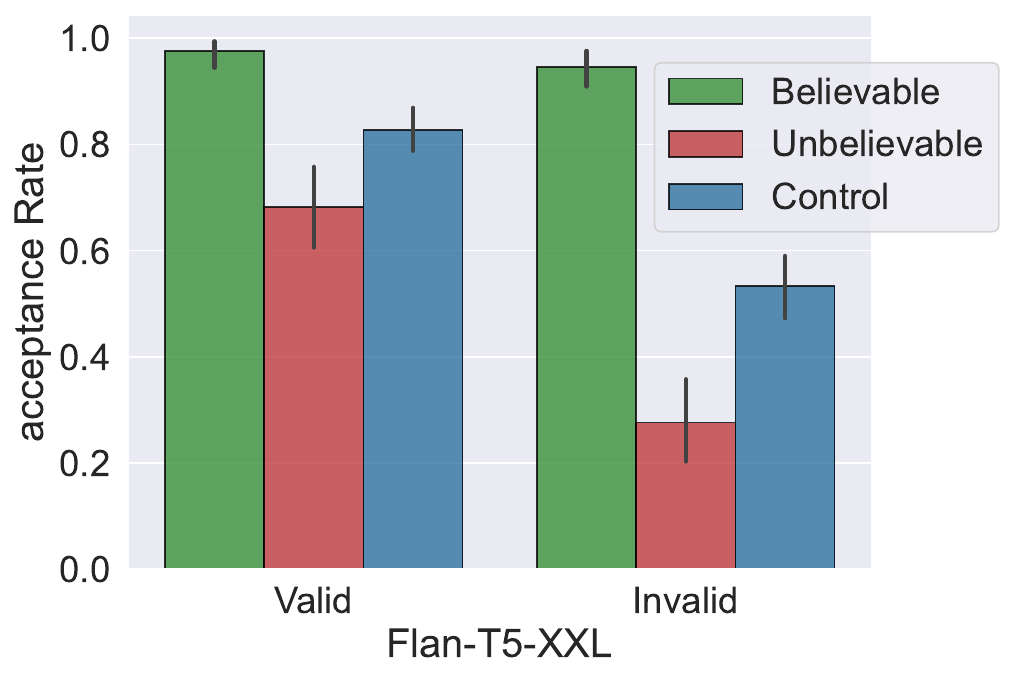}
\caption{Acceptance rates of the Flan-T5 models on believable (green) and unbelievable (red) arguments in the \biaseddataset condition and on neutral arguments in the control condition (blue) divided into valid and invalid arguments. The Belief Invalid bias score for the larger Flan-T5-XXL model (lower) seems lower compared to the smaller Flan-T5-XL (upper) because the model is less successful on the neutral arguments (blue).}
\label{fig:model_size_compare}
\end{figure}

\paragraph{GPT4 is also biased.}
The results comparing GPT4 to its predecessor in the GPT series are presented in Table \ref{tab:gpt4_results}.
Across our experiments, GPT4 demonstrates the highest bias score in the decoy expensive and decoy cheaper biases.
Although the bias scores are lower in the certainty, belief valid, and belief invalid biases, GPT4 still exhibits significant bias levels.

The decreased bias scores observed in belief biases might be attributed to the model training, at least partly aimed at enhancing logical reasoning.
Part of the GPT4 training data was designed to improve reasoning skills using data from MATH \cite{hendrycks2021measuring} and GSM-8K \cite{cobbe2021training}.
%\gabist{However, since GPT4 is entirely separate from previous models \gabis{wdym by this? separate from what models?} with possibly numerous other different characteristics, we cannot attribute the decreased bias scores to this specific change.}
However, since GPT4 might be different in many other ways from DaVinci-003, we cannot attribute the decreased bias scores to this specific change.
Beyond that, even with possibly improved reasoning, the model had less success mitigating bias in the decoy effect, which exhibited the most pronounced bias.
Furthermore, we encountered instances in the zero-shot setting where GPT4 refrained from providing explicit choices,
so we report one-shot results in the few-shot format as explained later in Section \ref{subsec:few_shot_results} (the zero-shot results when GPT4 did answer are similar to the one-shot results).
%and we provide further elaboration on this matter in Appendix \ref{appendix:gpt4_zero-shot}.

While GPT4 shows some mitigation of biases, the prominence of the decoy effect has increased, and all biases remain pronounced.
These findings suggest that biases remain relevant in models designed to address bias mitigation, such as GPT4 which was trained using RLHF to avoid social biases such as biases about sexuality and norms around marriage \cite{2303.08774}.

\paragraph{The effect of model size on bias emergence.}

Figure \ref{fig:model_size} shows the discrepancy in bias scores between the XL and XXL versions of  Flan-T5.
Consistent with prior research on social biases \cite{tal-etal-2022-fewer}, the larger XXL model exhibits higher bias scores for three bias types (decoy cheaper, certainly, and belief valid).
Surprisingly, the decoy expensive and belief invalid bias scores are lower for the XXL model, suggesting a presumable reduction in bias compared to the XL model.

% Surprisingly, the trend for the decoy expensive seemed to be reversed, with a higher score for the XL model, but this may results from a specific behavior of the XXL model, as discussed in Section \ref{subsec:decoy_analysis}.
% The belief invalid bias score is also lower for the XXL model, suggesting a reduction in bias compared to the XL model.

The reduction in belief invalid bias score could be attributed to the XXL model's lower accuracy in identifying invalid conclusions within the non-Real objects condition, as depicted in Figure \ref{fig:model_size_compare}.
Specifically, in the invalid-believable condition, the XXL model demonstrates a higher acceptance rate, indicating a greater presence of bias.
In contrast, in the invalid non-real objects condition, the XXL model displays a significantly elevated acceptance rate, leading to reduced overall accuracy and consequently lowering the bias score as per our defined calculation method (Section \ref{subsec:bias_score}).

As to the reduction of bias score in the decoy expensive, that may result from a specific behavior of the XXL model, as discussed in Section \ref{subsec:decoy_analysis}.

\section{Analysis}

We delve into the effects of few-shot in Section \ref{subsec:few_shot_results} and explore different attributes of the decoy effect and belief bias in Sections \ref{subsec:decoy_analysis} and \ref{subsec:belief_acc_bias}.
%Further analysis of the attributes of the certainty effect and additional decoy effect traits can be viewed in Appendix \ref{appendix:sec:additional_analysis}.
% This analysis aims to gain insights into the impact of different factors on bias scores and shed light on the behavior of the model.

\begin{figure}[t!]
\centering
\includegraphics[width=1\columnwidth]{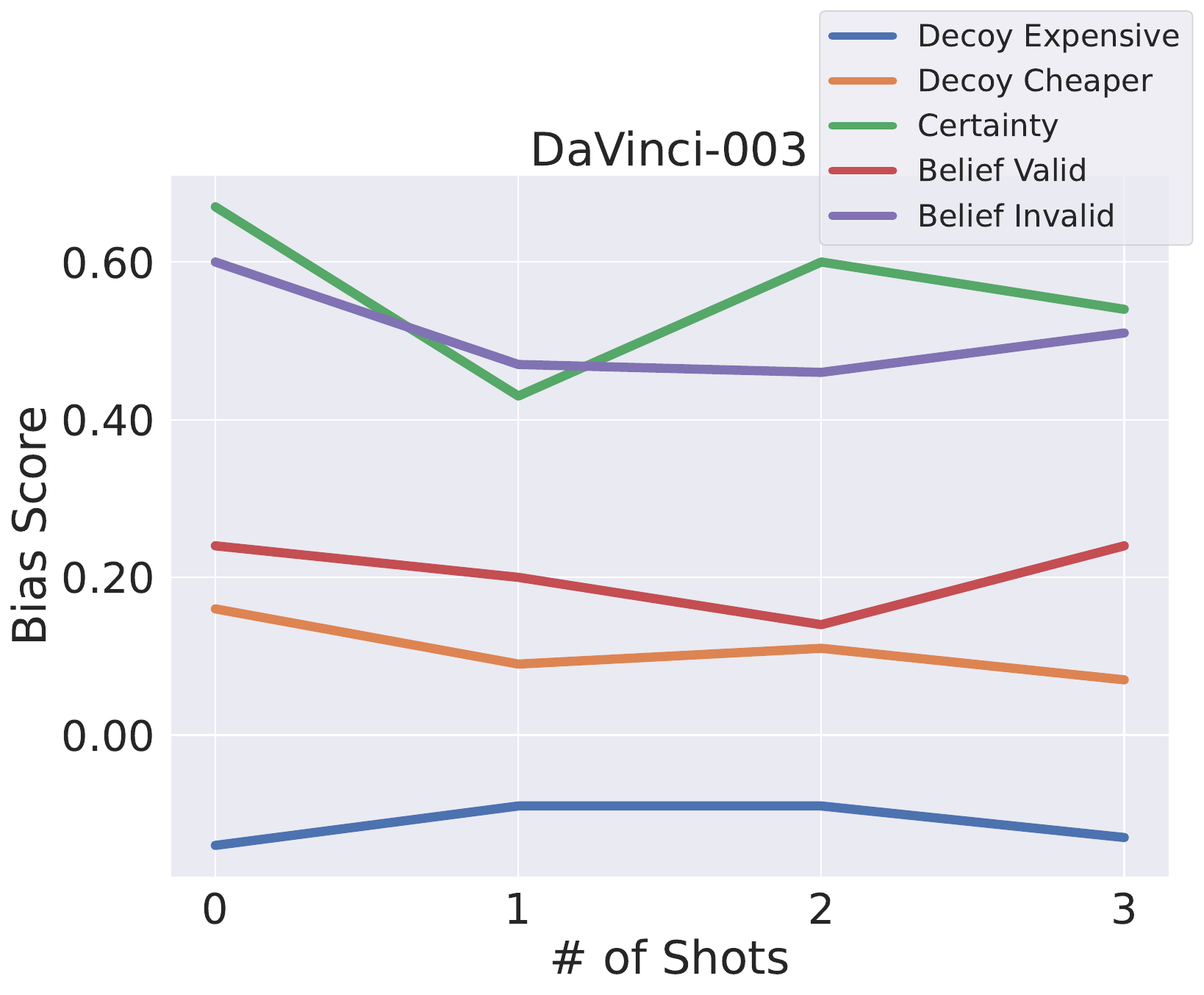}
% \qquad\qquad
%\includegraphics[width=1\columnwidth]{Sections/Figures/few_shot_text_davinci_003.pdf}
\includegraphics[width=1\columnwidth]{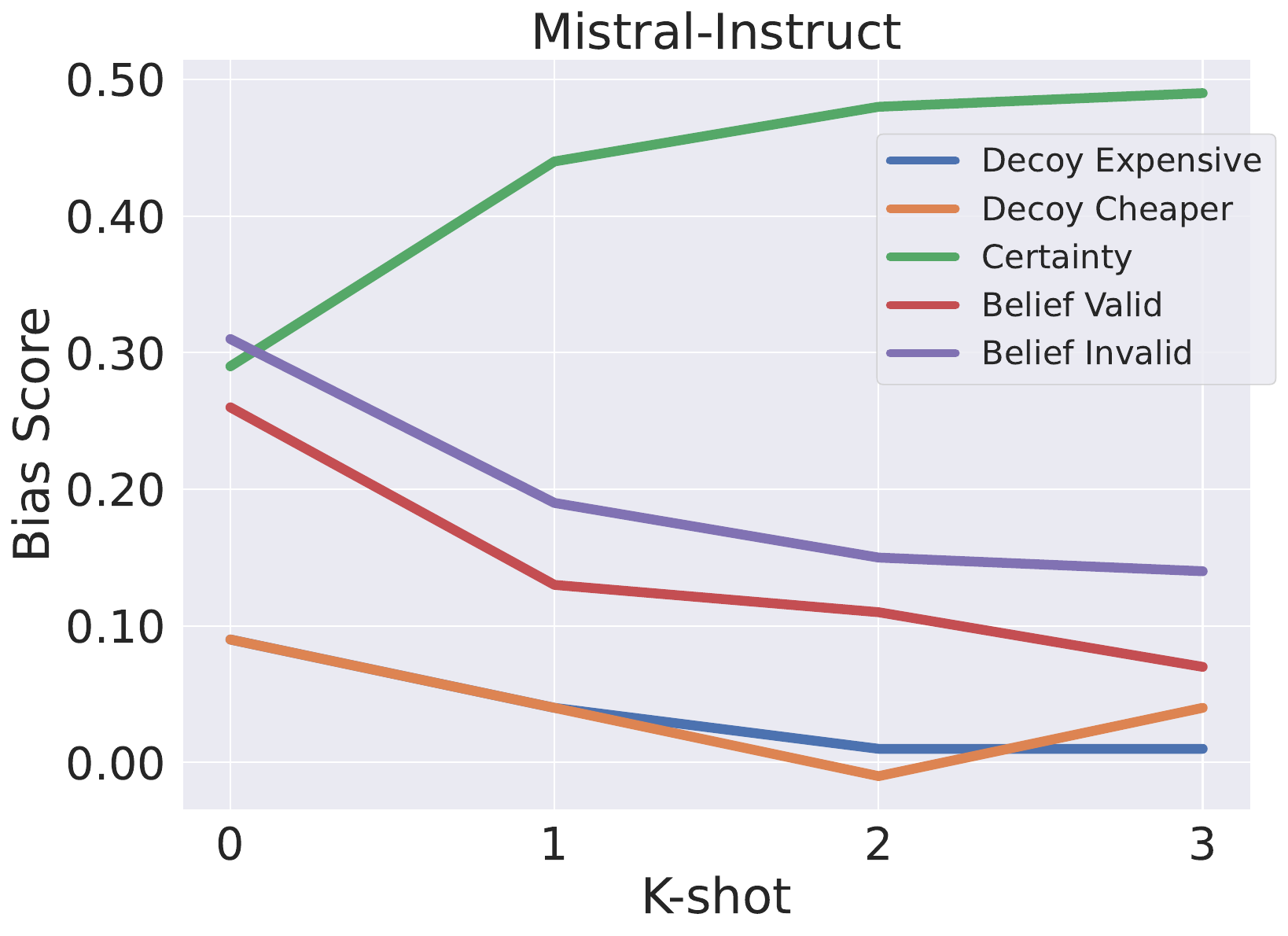}
\caption{The impact of format few-shots on bias scores using Davinci-003 \iedit{(top) and Mistral-Instruct (bottom)}. The utilization of few-shot examples \iedit{in most models} results in slightly lower bias scores\iedit{, while in Mistral-Instruct Belief biases are significantly lower and certainty bias increases}. To reduce computation costs, bias scores for Decoy Expensive and Decoy Cheaper biases are calculated solely on a specific product category (real-estate properties).}
\label{fig:format_few_shot}
\end{figure}

%\subsection{Few-shot}

\subsection{Using Few-shot}
\label{subsec:few_shot_results}
% % The decoy effect and certainty datasets used in our study are characterized by their choice-dependent nature, as they are decision-making biases wherein a clear and unequivocal correct answer does not exist.
% % Consequently, the zero-shot setting aligns most effectively with our objective of assessing the model's bias with minimal external influence.
% The samples used for the decoy and certainty effects are choice-dependent questions with no ``correct'' answer (recall the examples in Table 
% \ref{table:examples_biases}).
% It is therefore not obvious how to construct few-shot examples, which presumably should have correct labels in the prompt.
% Recent work suggested that giving few-shot samples without the correct labels could improve model performance by introducing the model with the overall format of
% the samples \cite{min2022rethinking}.
% However, in our case, giving task samples with random answers as few-shot examples could still affect the model since each answer is associated with either biased or unbiased behavior and could teach the model that this is the desired behavior type, thus compromising the results.

Our main experiments used a zero-shot setting to avoid giving the model examples that could bias it in any direction --- giving the model an example with an answer that is target could affect the tenacity of the model in choosing the target and vice versa.
Therefore, to help the model understand the sample format without biasing it in either direction, we experiment with few-shot prompting without the original samples.

Instead of using the samples from our datasets, we use manually curated choices between arbitrary options for the decoy and certainty effects (e.g., ``Which would you prefer, a white or black shirt?'') and mathematical reasoning examples for the belief bias (e.g., ``The price is \$10 per soda. The customer inserted 20\$. Conclusion: The customer can buy only 1 soda. Answer: Invalid.'').
We call this approach \textit{format few-shot} as the intention is to show the model the sample format using few-shot examples.
\iedit{We curated a 5-example pool for each bias, randomly selecting each sample from them.} 

In the case of the belief bias, there \emph{are} correct labels.
Therefore, we can also prompt the model with few-shot samples and avoid biasing the model by utilizing samples comprised of neutral non-real objects derived from a distinct set of fabricated words that were deliberately excluded from the test data.
We call this \textit{Task few-shot} as few-shot samples are from the same task as the test sample.
This approach enables us to assess the impact of using few-shot examples solely for formatting purposes compared to employing few-shot examples from the same task on the bias scores.

\begin{figure}[t!]
\centering
\includegraphics[width=1\columnwidth]{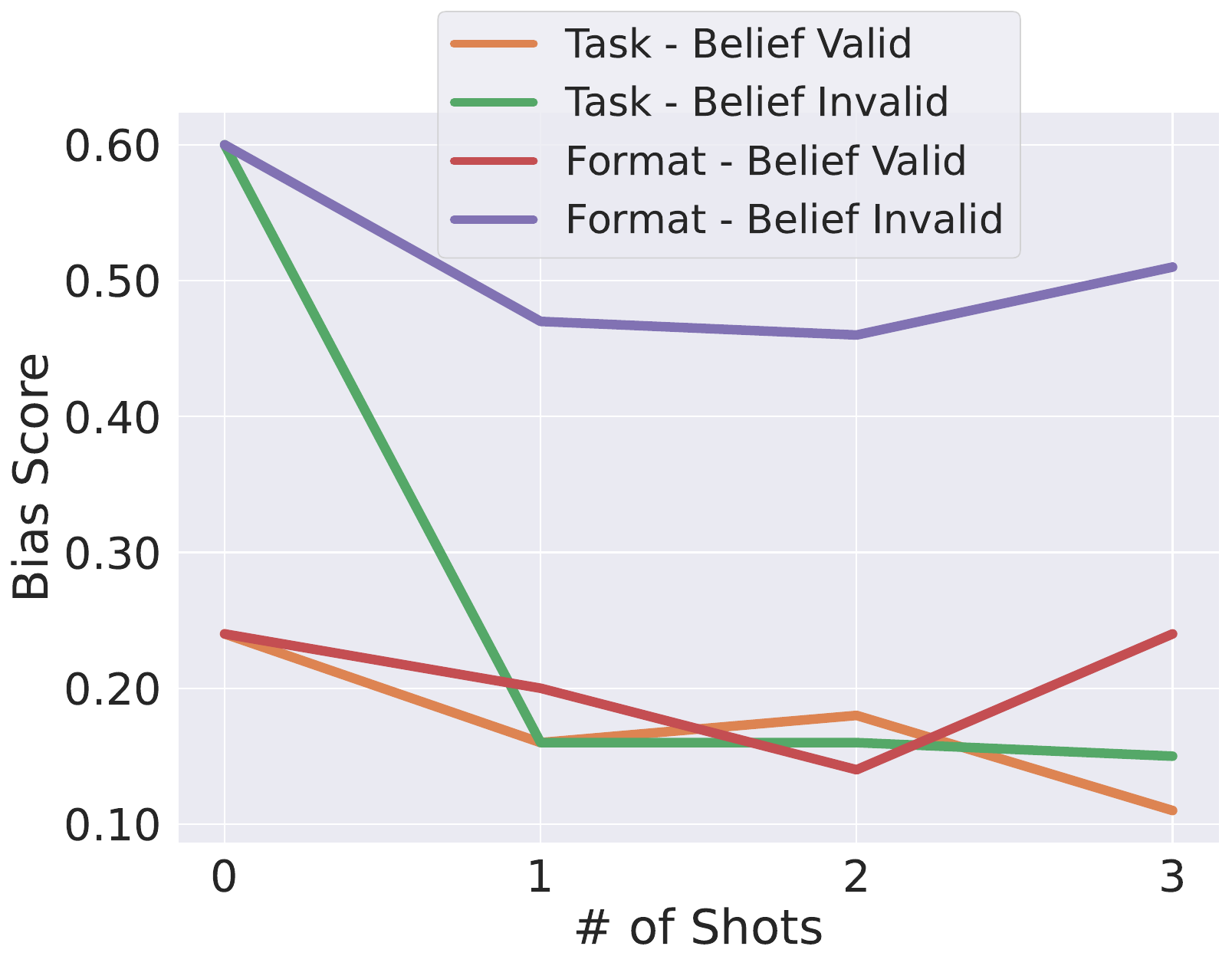}
% \qquad\qquad
\caption{The impact of format few-shots in comparison to task few-shots on bias scores, utilizing the DaVinci-003 model. When the model is prompted with examples from the same task, the decrease in bias scores is relatively lower compared to employing examples with merely the same format as the task.}
\label{fig:task_few_shot}
\end{figure}

\paragraph{Results.}
Results with format few-shot examples can be seen in Figure \ref{fig:format_few_shot}.
\ieditso{For}{Regarding DaVinci-003, in}  the decoy and certainty effects, there is no distinct trend when using the format few-shot examples, except for a small decrease in bias score when changing from zero-shot to one-shot setting.
Overall, increasing the number of few-shot examples might help the model understand the format but does not significantly decrease the bias.

In the case of the belief bias, incorporating few-shot examples leads to a noticeable reduction in the bias score, although a significant level of bias persists.
\iedit{This effect is more significant when utilizing task examples, as can be seen in Figure \ref{fig:task_few_shot}.}
This observation can perhaps be attributed to the presence of a logical reasoning process required by the belief bias examples, whereby the model's utilization of few-shot examples aids in facilitating problem-solving and helps to overcome the inherent bias associated with belief.

\iedit{While these results are similar to the other instruction models we tested, Mistral-Instruct demonstrated a unique behavior. Its belief and decoy bias scores consistently decreased, while the certainty bias score increased with additional format examples. Notably, the pretrained version of Mistral also observed a rise in the certainty effect bias score in the few-shot setting (increasing from 0.03 to 0.31). This exception entails we have much to learn about the effect of pertaining data and training techniques on the way models utilize few-shot in general and regarding biases specifically.}
\looseness=-1

This analysis focuses on the impact of few-shot examples solely on the instruction-tuned models, which exhibited the highest bias scores.
However, it is plausible to speculate that the pretrained models, which demonstrated the lowest bias scores, could potentially benefit even more significantly from learning the format through few-shot examples, considering their stronger dependence on understanding the format.
This could lead to the possible observation of higher bias scores for the pretrained models when giving few-shot samples.
To address this, we conducted few-shot experiments for the pretrained models, which revealed that the bias scores remain similarly low for these models with the exception of Mistral, as described before.
%To address this, we also include few-shot results for the pretrained models, revealing that the bias scores remain similarly low for these models.
%Refer to Appendix \ref{appendix:sec:few_shot_pretrained} for further details.

\subsection{Decoy Effect Analysis}
\label{subsec:decoy_analysis}
%- Detail about the different products' results
We investigate multiple attributes of the decoy effect, encompassing diverse product outcomes and price ranges, to assess their impact on the bias score and partly compare them with human behavior.
Moreover, we explore a particular behavior identified in the decoy expensive effect that has a notable influence on the bias scores.

\begin{figure}[t!]
\centering
\includegraphics[width=1\columnwidth]{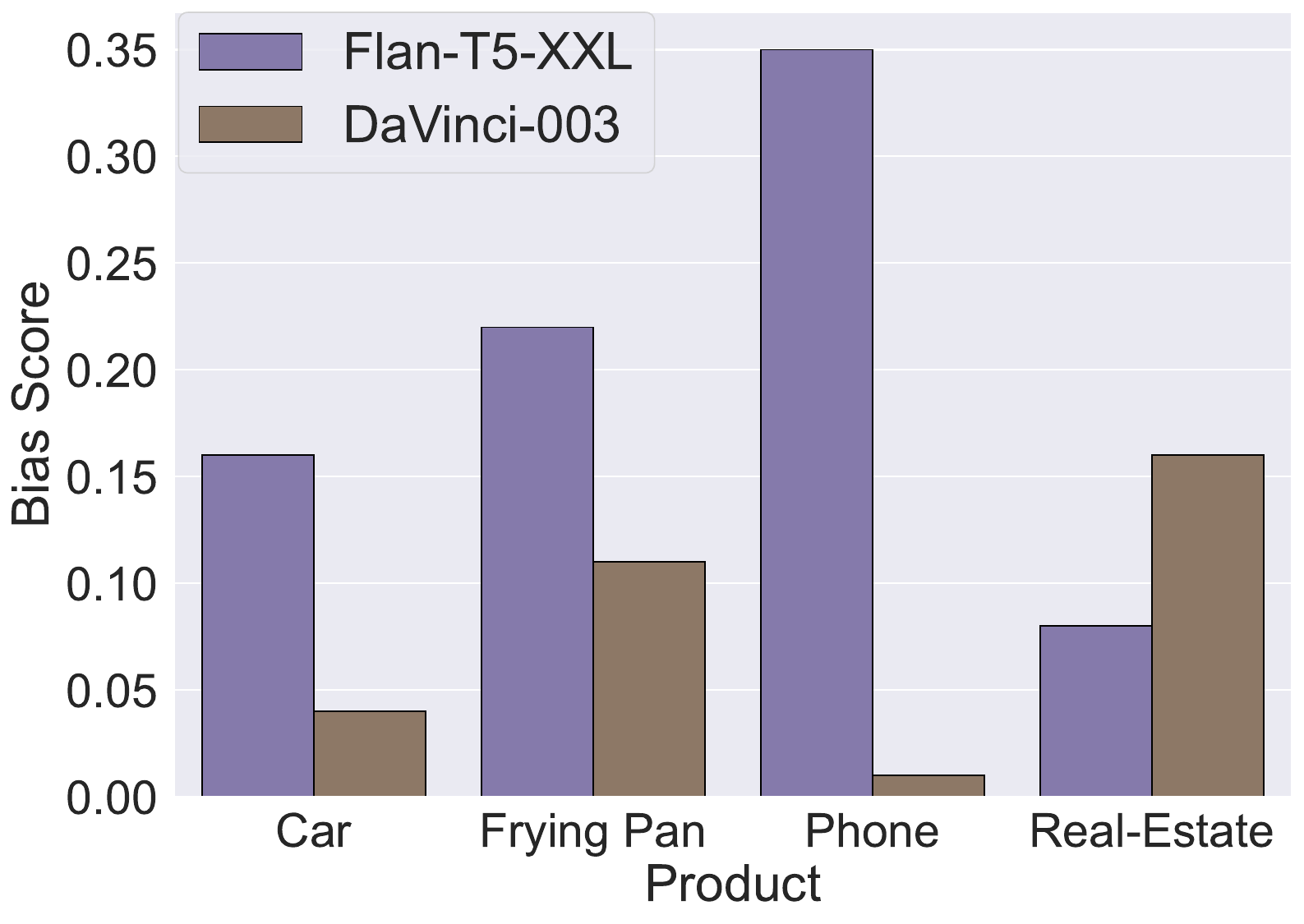}
% \qquad\qquad
\caption{
%\gabis{I'd remove the gray background in all figures}
The bias scores of the decoy \iedit{cheaper} effect across various products for the Flan-T5-XXL and DaVinci-003 models. The bias scores exhibit consistency \iedit{of bias existence} across all products, indicating that the observed behavior remains \iedit{more or less} uniform \iedit{within models} across different product categories and price ranges, akin to human \iedit{cognitive theory}.}
\label{fig:products}
\end{figure}

\paragraph{Products Variance.}
There was a moderate variance in bias scores in the decoy expensive results across different products, as shown in Figure \ref{fig:products}.

As the bias scores are computed as the difference between the \biaseddataset{} and control conditions, the score varies with the variance in the base preference of the model to target option in the control condition for each product.
Such differences between products were also observed to some extent in the original experiments on human subjects.

Together with the effect of price on the bias score (which is also analyzed in this section) and quality differences between products, the base preference of the model can cause a variance between different products.
%The full results for different products  can be viewed in Appendix \ref{appendix:subsec:products_results}

%- the way choices on the two options part set the limit for the bias scores.

% - \textbf{Show the effect of a more extreme decoy bias on a higher bias.}

\begin{figure}[t!]
\centering
\includegraphics[width=1\columnwidth]{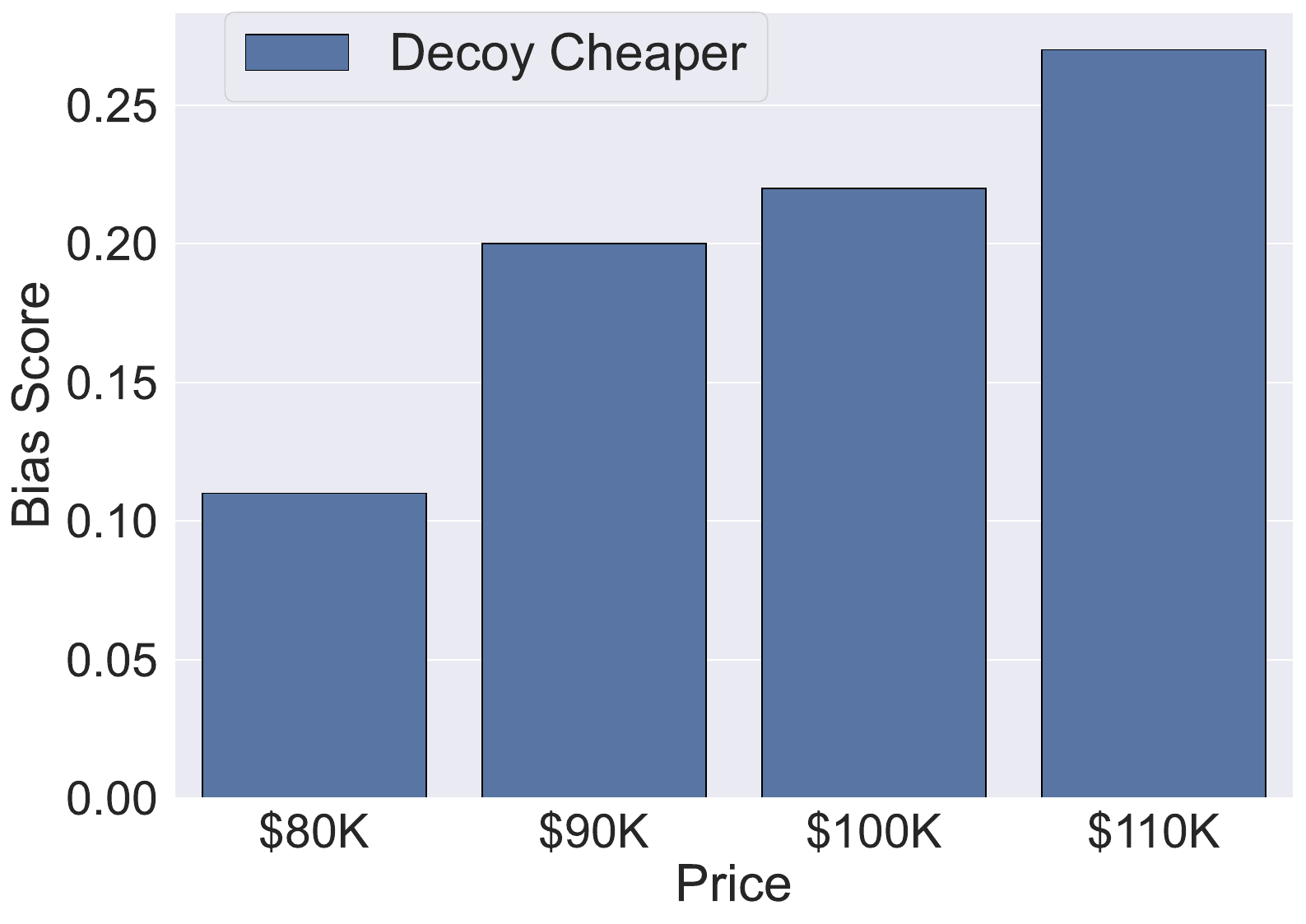}
% \qquad\qquad
\caption{The Effect of price range on the bias score of the decoy cheaper bias with real-estate products in the DaVinci-003 model. The x-axis represents the target price, where an increase in the target price leads to a wider gap between the target and competitor prices. The bias score demonstrates a positive correlation with the increasing price gap.}
\label{fig:decoy_prices}
\end{figure}

\paragraph{Price Range Effect.}

We investigate the relationship between the target price and the price gap, defined as the difference between the prices of the competitor option and the target option.
In our data, the higher the target value, the higher the gap from the competitor option.
By selecting values with varying price gaps, we aimed to examine the impact of this factor on bias scores.

As Figure \ref{fig:decoy_prices} shows, as the price range increases, the bias scores also exhibit higher values.
Although human experiments did not analyze this aspect, this result aligns with the expected behavior of this bias and is intuitively reasonable.

%Results for the Flan-T5 model were mostly similar, with one exception, as depicted in Appendix \ref{appendix:subsec:price_range}.

% \input{Sections/Figures/07_decoy_bias_type_flan}
% \paragraph{Decoy Sub-types.}
% As explained in Section \ref{subsec:def_decoy}, Given a target and a competitor target we generate four decoy options: (\textit{R}) higher price, same quality;
% (\textit{R*}) extremely higher price, same quality;
% (\textit{RF})  both higher price, worse quality; and (\textit{F})  same price, worse quality.

% The findings from the study conducted on human subjects \citep{huber1982adding} indicate that the price-increasing decoy sub-types (R and R*) exhibited the highest increase in bias score, followed by the RF sub-type with a relatively smaller gain.
% In comparison, the F sub-type demonstrated the smallest net gain in bias score.

% Figure \ref{fig:decoy_type_flan} presents the results for the Flan-T5 model.
% Intriguingly,  contrary to findings in human subjects, the F sub-type exhibits the highest increase in bias score, followed by a smaller increase in the R and R* sub-types.
% The DaVinci-003 model exhibited minimal variation in response to different decoy sub-types, as demonstrated in Appendix \ref{appendix:subsec:decoy_type_davinci003}.

% It is important to note that this comparison should be approached cautiously due to variations in data between human experiments and model experiments.
% Nevertheless, the findings suggest that while models demonstrate cognitive-like biases, their biases may manifest differently.

\paragraph{The Decoy Expensive Effect.}

A notable observation in the decoy expensive experiments is the significantly low bias score of --0.18 exhibited by the Flan-T5 XXL model.
We found that this score stems from the model consistently favoring the more expensive target option in the control condition with nearly 100\% preference. 

Considering the model's unwavering preference for the more target option in the absence of a decoy,  the addition of a decoy option cannot possibly shift its preference from the competitor option to the target option.
While this leaves no room for a bias score above zero, this preference for the more expensive option leads to negative results as the model picks the more expensive option even when adding a more expensive decoy option, leading to a shift from the target to the decoy option.

It is intriguing to observe such behaviors in these models that do not align with familiar cognitive biases but contradict basic human logic.
These findings necessitate further investigation that goes beyond cognitive-like biases before utilizing these models to aid in human decision-making.

\subsection{More Accurate and More Biased} \label{subsec:belief_acc_bias}

\begin{figure}[t!] 
\centering
\includegraphics[width=1\columnwidth]{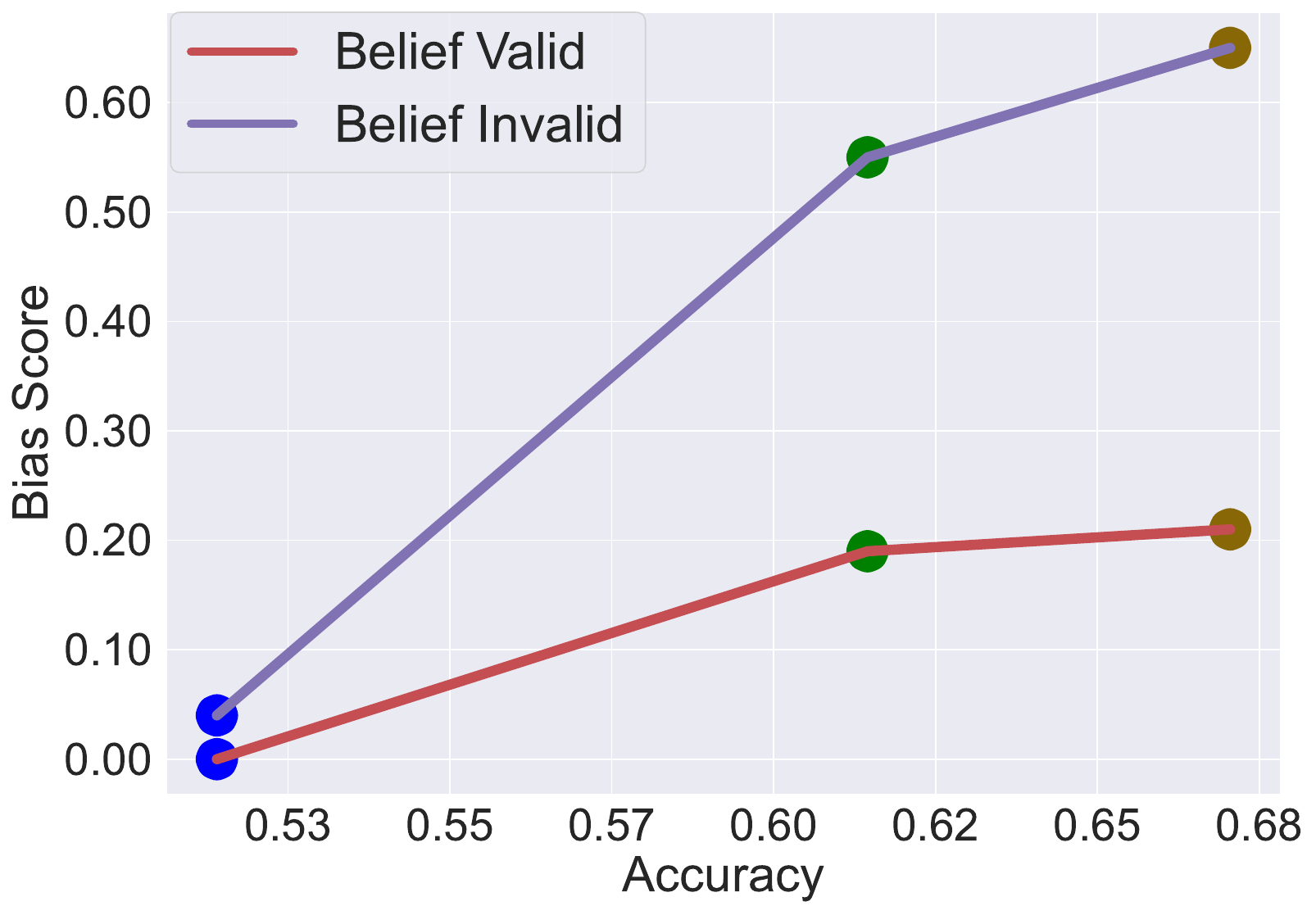}
% \qquad\qquad
\caption{The relationship between bias scores and model accuracy on the belief bias task's logical reasoning aspect for the DaVinci \iedit{ (blue)}, DaVinci-002 \iedit{ (green)}, and DaVinci-003 \iedit{ (brown)} models. Notably, an increase in model accuracy is accompanied by higher bias scores, indicating that improved accuracy does not necessarily mitigate biases in these models.}
\label{fig:belief_acc_bias_gpt}
\end{figure}

On the belief bias task, we can quantitatively measure the model accuracy, allowing us to examine the trade-off between accuracy and bias scores.

Figure \ref{fig:belief_acc_bias_gpt} shows the change in bias scores relative to the accuracy of the GPT models on the logical reasoning aspect of the belief bias task
%(see Appendix \ref{appendix:subsec:belief_acc_bias_t5} for  T5 results).
Interestingly, as the models demonstrate improved accuracy,  they also exhibit higher bias scores. This finding suggests that, despite advancements in accuracy, biases persist within these models.

Finally, our evaluation includes GPT4, a model specifically trained on logical reasoning.
GPT4 achieves a higher accuracy (84\%) compared to all GPT models, while simultaneously exhibiting a lower bias score than DaVinci-002 and DaVinci-003 (0.07 and 0.49 for the belief valid and belief invalid correspondingly).
This observation highlights the potential benefits of incorporating targeted training approaches to enhance both accuracy and mitigate biases in the process.

\section{Discussion and Conclusions
%\gabis{probably discussion and conclusion, or  discussion and future work?}
}

% Our investigation into cognitive biases in LMs fine-tuned on instructions and human feedback reveals their presence and prominence. We observe biases such as the decoy effect, the certainty effect, and the belief bias across these models.
% Despite efforts to mitigate biases, some still persist in LMs.
% This work contributes to understanding and addressing cognitive biases in instruction-tuned large LMs, aiming to develop more reliable and unbiased language models for practical applications.

Our study examines the influence of IT and RLHF on LMs' decision-making through cognitive bias analysis.
We reveal the presence of these biases across models, notably in models amplified with IT and RLHF.
These insights enhance our understanding of biases in fine-tuned models, widely considered superior to the pretrained models.

\iedit{In Section \ref{subsec:impact}, we explore the potential consequences of identifying these biases and the challenges in addressing them. Section \ref{subsec:origin_bias} delves into research paths investigating the source of these biases in the training of language models (LMs).}

\subsection{Real-World Impact}
\label{subsec:impact}

\iedit{The identified LM biases could impact real-world applications in decision-making and reasoning tasks.
For example, the presence of decoy and certainty effects 
may raise challenges for LMs as decision assistants.
Another impact could be reduced accuracy in some reasoning tasks in which the claim's plausibility plays a role.
This concern is demonstrated by the fact that in the belief bias, the treated samples exhibit a notable decrease in accuracy as expressed by the bias scores, ranging from 19\% to 61\%, compared to controlled samples.
%is that the belief bias treatment samples exhibit a notable decrease in accuracy, ranging from 18\% to 60\%, compared to control samples.
%This might indicate possible reduced accuracy in some cases in reasoning tasks such as common sense reasoning.
%with treatment samples showing a significant decrease in accuracy ranging from 60\% to 18\% compared to control samples.
Acknowledging and addressing these biases is crucial for enhancing the reliability and performance of LMs in real-world applications.}

One possible way to estimate the impact of these biases is to test models on biased datasets, before and after tuning. Nevertheless, using our proposed datasets for this process has its challenges, due to the complexities of controlling the amount of bias in models. For example, fine-tuning with belief bias control data might not reduce model bias, while using belief bias treatment data could improve logical reasoning but harm common sense. These complexities increase when considering effects like decoy and certainty, which lack defined truth labels. Although fine-tuning with our data is an appealing idea, it requires further investigation into how biases are learned in LLMs, which is beyond the scope of this paper.

\subsection{Origin of Bias}
\label{subsec:origin_bias}

\iedit{An additional question that arises concerns the origin of these biases.} Further research is needed to determine if biases come from pretraining, intensify during fine-tuning, or arise from a mix of both.
\iedit{While \citet{lin2023unlocking} claim that alignment methods only extract existing behavior models learned in pertaining, \citet{shwartz-choi-2020-neural} demonstrated that pretrained LMs tend to prioritize infrequent actions over more common ones, indicating the presence of reporting bias.
The biases outlined in our work may be associated with the prevalence of analogous questions and answers in the instruction and human feedback datasets used for model training.}
Studying bias-related examples in pretraining data and their magnification during fine-tuning can offer insights.
Evaluating the influence of different fine-tuning data and strategies on bias could illuminate fine-tuning's role in bias emergence.
Assessing how these biases interact with other known biases (such as reporting bias \cite{shwartz-choi-2020-neural}, and financial bias \cite{Zhou2024AreLL}) can provide insights into how they are acquired and potential interconnections.
Grasping these dynamics will guide strategies to improve model fairness and reliability.

\section{Limitations}

% This study has several limitations.
%First, the biases we examine are well-known, potentially leading to data contamination despite our efforts to introduce variations through different text templates and values we created ourselves.
%Second, the reliance on specific OpenAI models poses a limitation as their training details are undisclosed, and the future availability of these models is uncertain.
% Nonetheless, our findings contribute to the understanding of cognitive biases in language models and provide insights for bias mitigation in NLP systems.
%Beyond that, we focus only on English-based models so our results reflect existing biases only in English.
\iedit{In examining the impact of IT and RLHF on cognitive biases in LMs, our study highlights a notable challenge in disentangling the effects of different training datasets.
Flan-T5's IT data involves NLP tasks, Mistral-Instruct trained on unknown publicly available instructions datasets, while OpenAI's IT data uses assistant-like input-output pairs as far as we know.
The dissimilarity in training data makes it difficult to pinpoint the exact factors causing biases in our models, underscoring the need for further investigation.}

\iedit{Beyond that, the unavailability of information on OpenAI models' training limits our ability to draw clear conclusions.
Without details on their training procedures, we cannot determine whether RLHF training alone causes bias amplification or if GPT4's partial mitigation results from specific procedures, architecture differences, or other factors.
The uncertain future availability of OpenAI models puts the complete reproduction of the results at risk for future research.}
\iedit{Our study emphasizes the importance of transparency in model training for a better understanding of the relationship between IT and RLHF to biases in LMs.}

\iedit{Besides these model-specific limitations, there are limitations inherent in this type of research. One possible limitation is data contamination.
We address well-known biases that might leak into the training data despite our efforts to introduce new text and value variations.}

\iedit{
While it's common to evaluate pretrained LMs using answer probabilities \cite{NEURIPS2020_1457c0d6, holtzman-etal-2021-surface}, this evaluation method introduces a slight difference when compared to models trained on IT, which can be assessed based on their directly generated answers. Although unavoidable, this factor might influence results.}
We analyze the biases only in English-based models.

%\section{Future Work}
% Future work involves two key aspects. First, further exploration of both known and potentially unknown cognitive biases exhibited by language models is necessary to broaden our understanding.
% This will provide insights into the comprehensive landscape of biases in NLP systems.
% Second, it is crucial to advance debiasing methods based on promising initial findings. By refining and expanding these techniques, we can progress toward the development of more reliable and unbiased language models for practical applications.

%\section*{Acknowledgment}
%We would like to express our gratitude to %Ishita Dasgupta, Andrew K. Lampinen,and  
%\citet{dasgupta2022language} for generously sharing their valuable data.

\section*{Acknowledgements}
This research was supported by the Israel Science Foundation (grants 448/20 and 278/22), an AI alignment grant from Open Philanthropy, and an Azrieli Foundation Early Career Faculty Fellowship.
\looseness=-1

\bibliography{tacl2021,custom}

\begin{thebibliography}{30}
\expandafter\ifx\csname natexlab\endcsname\relax\def\natexlab#1{#1}\fi

\bibitem[{Acciarini et~al.(2021)Acciarini, Brunetta, and
  Boccardelli}]{acciarini2021cognitive}
Chiara Acciarini, Federica Brunetta, and Paolo Boccardelli. 2021.
\newblock Cognitive biases and decision-making strategies in times of change: a
  systematic literature review.
\newblock \emph{Management Decision}, 59(3):638--652.

\bibitem[{Bai et~al.(2022)Bai, Jones, Ndousse, Askell, Chen, DasSarma, Drain,
  Fort, Ganguli, Henighan et~al.}]{bai2022training}
Yuntao Bai, Andy Jones, Kamal Ndousse, Amanda Askell, Anna Chen, Nova DasSarma,
  Dawn Drain, Stanislav Fort, Deep Ganguli, Tom Henighan, et~al. 2022.
\newblock Training a helpful and harmless assistant with reinforcement learning
  from human feedback.
\newblock \emph{arXiv preprint arXiv:2204.05862}.

\bibitem[{Berthet(2022)}]{berthet2022impact}
Vincent Berthet. 2022.
\newblock The impact of cognitive biases on professionals’ decision-making: A
  review of four occupational areas.
\newblock \emph{Frontiers in Psychology}, 12:802439.

\bibitem[{Binz and Schulz(2022)}]{Binz2022UsingCP}
Marcel Binz and Eric Schulz. 2022.
\newblock Using cognitive psychology to understand gpt-3.
\newblock \emph{ArXiv}, abs/2206.14576.

\bibitem[{Brown et~al.(2020)Brown, Mann, Ryder, Subbiah, Kaplan, Dhariwal,
  Neelakantan, Shyam, Sastry, Askell, Agarwal, Herbert-Voss, Krueger, Henighan,
  Child, Ramesh, Ziegler, Wu, Winter, Hesse, Chen, Sigler, Litwin, Gray, Chess,
  Clark, Berner, McCandlish, Radford, Sutskever, and
  Amodei}]{NEURIPS2020_1457c0d6}
Tom Brown, Benjamin Mann, Nick Ryder, Melanie Subbiah, Jared~D Kaplan, Prafulla
  Dhariwal, Arvind Neelakantan, Pranav Shyam, Girish Sastry, Amanda Askell,
  Sandhini Agarwal, Ariel Herbert-Voss, Gretchen Krueger, Tom Henighan, Rewon
  Child, Aditya Ramesh, Daniel Ziegler, Jeffrey Wu, Clemens Winter, Chris
  Hesse, Mark Chen, Eric Sigler, Mateusz Litwin, Scott Gray, Benjamin Chess,
  Jack Clark, Christopher Berner, Sam McCandlish, Alec Radford, Ilya Sutskever,
  and Dario Amodei. 2020.
\newblock \href
  {https://proceedings.neurips.cc/paper/2020/file/1457c0d6bfcb4967418bfb8ac142f64a-Paper.pdf}
  {Language models are few-shot learners}.
\newblock In \emph{Advances in Neural Information Processing Systems},
  volume~33, pages 1877--1901. Curran Associates, Inc.

\bibitem[{Chung et~al.(2022)Chung, Hou, Longpre, Zoph, Tay, Fedus, Li, Wang,
  Dehghani, Brahma et~al.}]{chung2022scaling}
Hyung~Won Chung, Le~Hou, Shayne Longpre, Barret Zoph, Yi~Tay, William Fedus,
  Eric Li, Xuezhi Wang, Mostafa Dehghani, Siddhartha Brahma, et~al. 2022.
\newblock Scaling instruction-finetuned language models.
\newblock \emph{arXiv preprint arXiv:2210.11416}.

\bibitem[{Cobbe et~al.(2021)Cobbe, Kosaraju, Bavarian, Chen, Jun, Kaiser,
  Plappert, Tworek, Hilton, Nakano et~al.}]{cobbe2021training}
Karl Cobbe, Vineet Kosaraju, Mohammad Bavarian, Mark Chen, Heewoo Jun, Lukasz
  Kaiser, Matthias Plappert, Jerry Tworek, Jacob Hilton, Reiichiro Nakano,
  et~al. 2021.
\newblock Training verifiers to solve math word problems.
\newblock \emph{arXiv preprint arXiv:2110.14168}.

\bibitem[{Dasgupta et~al.(2022)Dasgupta, Lampinen, Chan, Creswell, Kumaran,
  McClelland, and Hill}]{dasgupta2022language}
Ishita Dasgupta, Andrew~K Lampinen, Stephanie~CY Chan, Antonia Creswell,
  Dharshan Kumaran, James~L McClelland, and Felix Hill. 2022.
\newblock Language models show human-like content effects on reasoning.
\newblock \emph{arXiv preprint arXiv:2207.07051}.

\bibitem[{Dimick and Ryan(2014)}]{dimick2014methods}
Justin~B Dimick and Andrew~M Ryan. 2014.
\newblock Methods for evaluating changes in health care policy: the
  difference-in-differences approach.
\newblock \emph{Jama}, 312(22):2401--2402.

\bibitem[{Evans et~al.(1983)Evans, Barston, and Pollard}]{evans1983conflict}
JSBT Evans, Julie~L Barston, and Paul Pollard. 1983.
\newblock On the conflict between logic and belief in syllogistic reasoning.
\newblock \emph{Memory \& cognition}, 11(3):295--306.

\bibitem[{Friedman and Savage(1948)}]{friedman1948utility}
Milton Friedman and Leonard~J Savage. 1948.
\newblock The utility analysis of choices involving risk.
\newblock \emph{Journal of political Economy}, 56(4):279--304.

\bibitem[{Gonen and Goldberg(2019)}]{gonen2019lipstick}
Hila Gonen and Yoav Goldberg. 2019.
\newblock Lipstick on a pig: Debiasing methods cover up systematic gender
  biases in word embeddings but do not remove them.
\newblock \emph{arXiv preprint arXiv:1903.03862}.

\bibitem[{Hagendorff et~al.(2022)Hagendorff, Fabi, and
  Kosinski}]{hagendorff2022machine}
Thilo Hagendorff, Sarah Fabi, and Michal Kosinski. 2022.
\newblock Machine intuition: Uncovering human-like intuitive decision-making in
  gpt-3.5.
\newblock \emph{arXiv preprint arXiv:2212.05206}.

\bibitem[{Hendrycks et~al.(2021)Hendrycks, Burns, Kadavath, Arora, Basart,
  Tang, Song, and Steinhardt}]{hendrycks2021measuring}
Dan Hendrycks, Collin Burns, Saurav Kadavath, Akul Arora, Steven Basart, Eric
  Tang, Dawn Song, and Jacob Steinhardt. 2021.
\newblock Measuring mathematical problem solving with the math dataset.
\newblock \emph{arXiv preprint arXiv:2103.03874}.

\bibitem[{Holtzman et~al.(2021)Holtzman, West, Shwartz, Choi, and
  Zettlemoyer}]{holtzman-etal-2021-surface}
Ari Holtzman, Peter West, Vered Shwartz, Yejin Choi, and Luke Zettlemoyer.
  2021.
\newblock \href {https://doi.org/10.18653/v1/2021.emnlp-main.564} {Surface form
  competition: Why the highest probability answer isn{'}t always right}.
\newblock In \emph{Proceedings of the 2021 Conference on Empirical Methods in
  Natural Language Processing}, pages 7038--7051, Online and Punta Cana,
  Dominican Republic. Association for Computational Linguistics.

\bibitem[{Huber et~al.(1982)Huber, Payne, and Puto}]{huber1982adding}
Joel Huber, John~W Payne, and Christopher Puto. 1982.
\newblock Adding asymmetrically dominated alternatives: Violations of
  regularity and the similarity hypothesis.
\newblock \emph{Journal of consumer research}, 9(1):90--98.

\bibitem[{Jiang et~al.(2023)Jiang, Sablayrolles, Mensch, Bamford, Chaplot,
  Casas, Bressand, Lengyel, Lample, Saulnier et~al.}]{jiang2023mistral}
Albert~Q Jiang, Alexandre Sablayrolles, Arthur Mensch, Chris Bamford,
  Devendra~Singh Chaplot, Diego de~las Casas, Florian Bressand, Gianna Lengyel,
  Guillaume Lample, Lucile Saulnier, et~al. 2023.
\newblock Mistral 7b.
\newblock \emph{arXiv preprint arXiv:2310.06825}.

\bibitem[{Kahneman(1979)}]{kahneman1979prospect}
Daniel Kahneman. 1979.
\newblock Prospect theory: An analysis of decisions under risk.
\newblock \emph{Econometrica}, 47:278.

\bibitem[{Lin et~al.(2023)Lin, Ravichander, Lu, Dziri, Sclar, Chandu,
  Bhagavatula, and Choi}]{lin2023unlocking}
Bill~Yuchen Lin, Abhilasha Ravichander, Ximing Lu, Nouha Dziri, Melanie Sclar,
  Khyathi Chandu, Chandra Bhagavatula, and Yejin Choi. 2023.
\newblock The unlocking spell on base llms: Rethinking alignment via in-context
  learning.
\newblock \emph{arXiv preprint arXiv:2312.01552}.

\bibitem[{McFadden(1974)}]{mcfadden1974conditional}
D~McFadden. 1974.
\newblock Conditional logit analysis of qualitative choice behavior.
\newblock \emph{Frontiers in Econometrics}.

\bibitem[{Min et~al.(2022)Min, Lyu, Holtzman, Artetxe, Lewis, Hajishirzi, and
  Zettlemoyer}]{min2022rethinking}
Sewon Min, Xinxi Lyu, Ari Holtzman, Mikel Artetxe, Mike Lewis, Hannaneh
  Hajishirzi, and Luke Zettlemoyer. 2022.
\newblock Rethinking the role of demonstrations: What makes in-context learning
  work?
\newblock \emph{arXiv preprint arXiv:2202.12837}.

\bibitem[{OpenAI(2023)}]{2303.08774}
OpenAI. 2023.
\newblock \href {http://arxiv.org/abs/arXiv:2303.08774} {Gpt-4 technical
  report}.

\bibitem[{Ouyang et~al.(2022)Ouyang, Wu, Jiang, Almeida, Wainwright, Mishkin,
  Zhang, Agarwal, Slama, Ray, Schulman, Hilton, Kelton, Miller, Simens, Askell,
  Welinder, Christiano, Leike, and Lowe}]{Ouyang2022TrainingLM}
Long Ouyang, Jeff Wu, Xu~Jiang, Diogo Almeida, Carroll~L. Wainwright, Pamela
  Mishkin, Chong Zhang, Sandhini Agarwal, Katarina Slama, Alex Ray, John
  Schulman, Jacob Hilton, Fraser Kelton, Luke~E. Miller, Maddie Simens, Amanda
  Askell, Peter Welinder, Paul~Francis Christiano, Jan Leike, and Ryan~J. Lowe.
  2022.
\newblock Training language models to follow instructions with human feedback.
\newblock \emph{ArXiv}, abs/2203.02155.

\bibitem[{Raffel et~al.(2020)Raffel, Shazeer, Roberts, Lee, Narang, Matena,
  Zhou, Li, and Liu}]{raffel2020exploring}
Colin Raffel, Noam Shazeer, Adam Roberts, Katherine Lee, Sharan Narang, Michael
  Matena, Yanqi Zhou, Wei Li, and Peter~J Liu. 2020.
\newblock Exploring the limits of transfer learning with a unified text-to-text
  transformer.
\newblock \emph{The Journal of Machine Learning Research}, 21(1):5485--5551.

\bibitem[{Shwartz and Choi(2020)}]{shwartz-choi-2020-neural}
Vered Shwartz and Yejin Choi. 2020.
\newblock \href {https://doi.org/10.18653/v1/2020.coling-main.605} {Do neural
  language models overcome reporting bias?}
\newblock In \emph{Proceedings of the 28th International Conference on
  Computational Linguistics}, pages 6863--6870, Barcelona, Spain (Online).
  International Committee on Computational Linguistics.

\bibitem[{Smith(2022)}]{sep-aristotle-logic}
Robin Smith. 2022.
\newblock {Aristotle’s Logic}.
\newblock In Edward~N. Zalta and Uri Nodelman, editors, \emph{The {Stanford}
  Encyclopedia of Philosophy}, {W}inter 2022 edition. Metaphysics Research Lab,
  Stanford University.

\bibitem[{Student(1908)}]{student1908probable}
Student. 1908.
\newblock The probable error of a mean.
\newblock \emph{Biometrika}, 6(1):1--25.

\bibitem[{Tal et~al.(2022)Tal, Magar, and Schwartz}]{tal-etal-2022-fewer}
Yarden Tal, Inbal Magar, and Roy Schwartz. 2022.
\newblock \href {https://doi.org/10.18653/v1/2022.gebnlp-1.13} {Fewer errors,
  but more stereotypes? the effect of model size on gender bias}.
\newblock In \emph{Proceedings of the 4th Workshop on Gender Bias in Natural
  Language Processing (GeBNLP)}, pages 112--120, Seattle, Washington.
  Association for Computational Linguistics.

\bibitem[{Zhou et~al.(2023)Zhou, Li, Li, Yu, Liu, Wang, Zhang, Ji, Yan, He
  et~al.}]{zhou2023comprehensive}
Ce~Zhou, Qian Li, Chen Li, Jun Yu, Yixin Liu, Guangjing Wang, Kai Zhang, Cheng
  Ji, Qiben Yan, Lifang He, et~al. 2023.
\newblock A comprehensive survey on pretrained foundation models: A history
  from bert to chatgpt.
\newblock \emph{arXiv preprint arXiv:2302.09419}.

\bibitem[{Zhou et~al.(2024)Zhou, Ni, Liu, Zhang, Liu, Ye, and
  Chai}]{Zhou2024AreLL}
Yuhang Zhou, Yuchen Ni, Xiang Liu, Jian Zhang, Sen Liu, Guangnan Ye, and
  Hongfeng Chai. 2024.
\newblock \href {https://api.semanticscholar.org/CorpusID:267760135} {Are large
  language models rational investors?}

\end{thebibliography}
\bibliographystyle{acl_natbib}

\iftaclpubformat

\onecolumn

\appendix
\fi

\end{document}